\documentclass{article}

     \usepackage[final,nonatbib]{neurips_2023}

\usepackage[utf8]{inputenc} % allow utf-8 input
\usepackage[T1]{fontenc}    % use 8-bit T1 fonts
\usepackage{url}            % simple URL typesetting
\usepackage{booktabs}       % professional-quality tables
\usepackage{amsfonts}       % blackboard math symbols
\usepackage{nicefrac}       % compact symbols for 1/2, etc.
\usepackage{microtype}      % microtypography
\usepackage{times}
\usepackage{epsfig}
\usepackage{graphicx}
\usepackage{amsmath}
\usepackage{amsthm}           % for proof
\usepackage{amssymb}
\usepackage{array}
\usepackage{multirow}
\usepackage[font=small,labelfont=bf]{caption}
\usepackage{cite}
\usepackage{enumitem}            % for noindent itemize
\usepackage{wasysym}
\usepackage{array, makecell} %
\usepackage{caption}
\usepackage{adjustbox}
\usepackage[table]{xcolor}       % for colored stuff
\usepackage{makecell}            % For thicker horizontal rules in tables

\usepackage{hhline}
\usepackage{nicematrix}
\usepackage{xr-hyper}
\usepackage{xspace}              % for dynamic controlled space at end of abbreviation

\newtheorem{theorem}{Theorem}

\newtheorem{lemma}{Lemma}

\newcommand{\methodName}{PrObeD\xspace}

\newcommand{\iou}{IoU\xspace}
\newcommand{\image}{\vect{I}_j}
\newcommand{\step}{s}
\newcommand{\stepSum}{\mathcal{S}}
\newcommand{\instant}{t}
\newcommand{\norm}[1]{\left\lVert#1\right\rVert}

\newcommand{\NCFourK}{NC$4$K\xspace}
\newcommand{\CODTenK}{COD$10$K\xspace}

\usepackage{pifont}% http://ctan.org/pkg/pifont
\newcommand{\cmark}{\checkmark}
\newcommand{\xmark}{\ding{53}}

\newcommand{\forExample}{\textit{e.g.}\xspace}
\newcommand{\thatIs}{\textit{i.e.}\xspace}

\newcommand{\etal}{\textit{et~al.}\xspace}

\providecommand\rightarrowRHD{\relbar\joinrel\mathrel\RHD}
\newcommand{\myTopRule}{\Xhline{2\arrayrulewidth}}

\usepackage[pagebackref,breaklinks,colorlinks,citecolor=orange,urlcolor=black]{hyperref}      % hyperlinks
\usepackage[capitalize]{cleveref}
\crefname{section}{Sec.}{Secs.}
\Crefname{section}{Section}{Sections}
\Crefname{table}{Table}{Tables}
\crefname{table}{Tab.}{Tabs.}

\definecolor{mygray}{gray}{0.95}
\newcommand{\vect}[1]{\boldsymbol{#1}}
\newcommand{\minisection}[1]{\vspace{1mm} \noindent \textbf{#1} \hspace{0mm}}
\usepackage{xcolor,pifont}
\newcommand*\colourcheck[1]{%
  \expandafter\newcommand\csname #1check\endcsname{\textcolor{#1}{\ding{52}}}%
}

\newcommand*\colourcross[1]{%
  \expandafter\newcommand\csname #1check\endcsname{\textcolor{#1}{\ding{54}}}%
}

\colourcheck{green}
\colourcross{red}

\setlist[itemize]{noitemsep,topsep=0pt,leftmargin=*,label={\large\textbullet}}

\newcommand{\toptitlebar}{
  \hrule height 0.05in
  \vskip 0.15in
  \vskip -\parskip%
}
\newcommand{\bottomtitlebar}{
  \vskip 0.2in
  \vskip -\parskip
  \hrule height 0.01in
  \vskip 0.09in%
}

\def\ie{\emph{i.e}.} 
 
 \def\vs{\emph{vs}.}
\def\wrt{w.r.t.} 
 
\def\etal{\emph{et al}.}

\title{\textcolor{red}{\methodName}: \textcolor{red}{Pr}oactive \textcolor{red}{Ob}j\textcolor{red}{e}ct \textcolor{red}{D}etection Wrapper}

\author{%
  Vishal Asnani\\
  Michigan State University\\
  \texttt{asnanivi@msu.edu} \\
  \And
  Abhinav Kumar\\
  Michigan State University\\
  \texttt{kumarab6@msu.edu} \\
  \AND
  Suya You\\
  DEVCOM Army Research Laboratory\\
  \texttt{suya.you.civ@army.mil} \\
  \And 
  Xiaoming Liu\\
  Michigan State University\\
  \texttt{liuxm@cse.msu.edu}
}
\begin{document}

\maketitle

%============================================================================
%============================================================================
%============================================================================
\begin{abstract}
Previous research in $2D$ object detection focuses on various tasks, including detecting objects in generic and camouflaged images. These works are regarded as passive works for object detection as they take the input image as is. However, convergence to global minima is not guaranteed to be optimal in neural networks; therefore, we argue that the trained weights in the object detector are not optimal. To rectify this problem, we propose a wrapper based on proactive schemes, PrObeD, which enhances the performance of these object detectors by learning a signal. PrObeD consists of an encoder-decoder architecture, where the encoder network generates an image-dependent signal termed templates to encrypt the input images, and the decoder recovers this template from the encrypted images. We propose that learning the optimum template results in an object detector with an improved detection performance. The template acts as a mask to the input images to highlight semantics useful for the object detector. Finetuning the object detector with these encrypted images enhances the detection performance for both generic and camouflaged. Our experiments on MS-COCO, CAMO, \CODTenK, and \NCFourK datasets show improvement over different detectors after applying PrObeD. Our models/codes are available at \url{https://github.com/vishal3477/Proactive-Object-Detection}. 
\end{abstract}

%============================================================================
%============================================================================
%============================================================================
\section{Introduction}
Generic $2D$ object detection (GOD) has improved from earlier traditional detectors~\cite{viola2001rapid, viola2004robust, dalal2005histograms, felzenszwalb2008discriminatively} to the deep-learning-based object detectors~\cite{ren2015faster, redmon2016you, chen2022diffusiondet, carion2020end, gupta2019lvis, he2019bounding}. 
Advancements in deep-learning-based methods underwent many architectural change over recent years, including one-stage~\cite{redmon2016you, redmon2018yolov3, bochkovskiy2020yolov4, redmon2017yolo9000, liu2016ssd, lin2017focal}, two-stage~\cite{girshick2014rich,girshick2015fast, ren2015faster}, CNN-based~\cite{girshick2015fast, redmon2016you,redmon2018yolov3, bochkovskiy2020yolov4, derakhshani2019assisted,fidler2013bottom,gidaris2015object, dai2016r}, transformer-based~\cite{carion2020end, zhu2020deformable}, and diffusion-based~\cite{chen2022diffusiondet} methods. 
All these methods aim to predict the $2D$ bounding box of the objects in the images and their category class. 

Another emerging area related to generic object detection is camouflaged object detection~\cite{fan2020camouflaged, fan2021concealed, ji2023deep, li2021uncertainty, he2023camouflaged, he2023strategic, he2023weakly} (COD).
COD aims to detect and segment objects blended with the background~\cite{fan2020camouflaged, fan2021concealed} via object-level mask supervision. 
Applications of COD include medical~\cite{fan2020pranet, liu2021covid}, surveillance~\cite{chen2022pedestrian} and autonomous driving~\cite{xue2018survey}. 
Early COD detectors exploit hand-crafted features~\cite{sengottuvelan2008performance, pan2011study} and optical flow~\cite{hou2011detection}, while current methods are deep-learning-based.
These methods utilize attention~\cite{sun2021context, chen2022camouflaged}, joint learning~\cite{li2021uncertainty}, image gradient~\cite{ji2023deep}, and transformers~\cite{mao2021transformer, yang2021uncertainty}. 

\begin{figure}[t!]
\centering
\includegraphics[trim={0 -4 0 0},clip,width=1\textwidth]{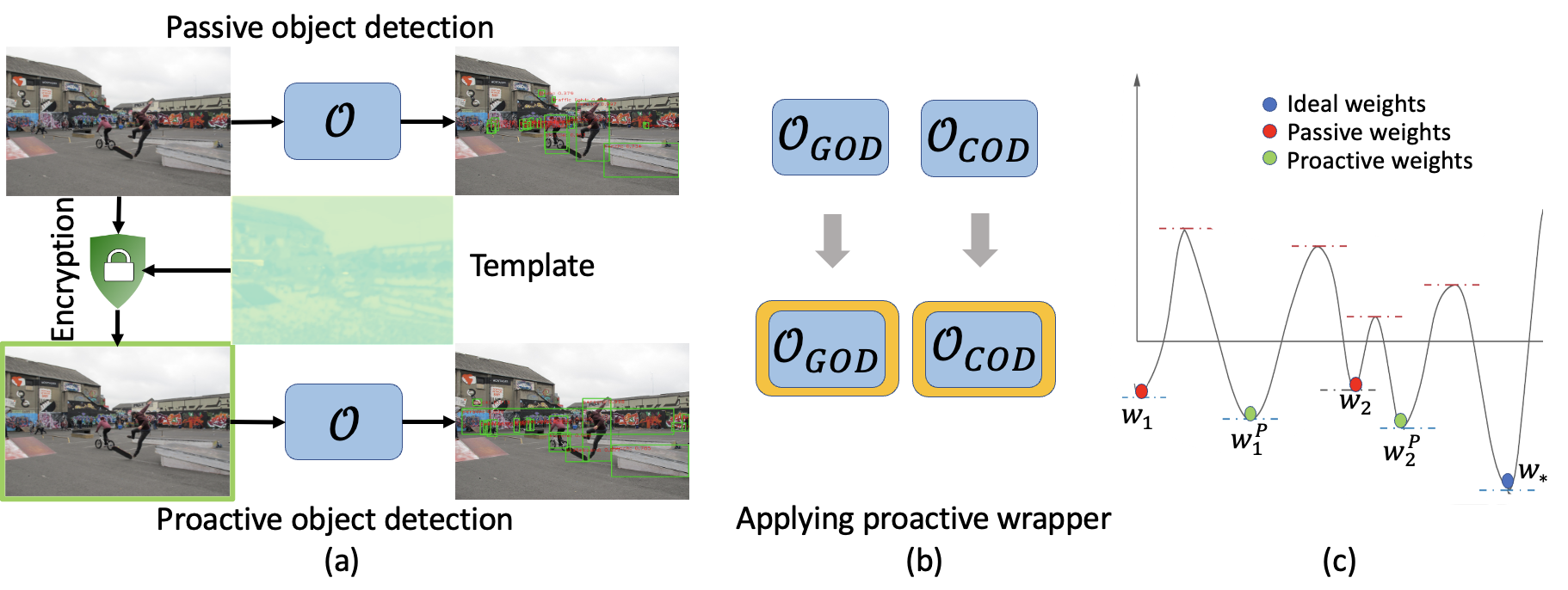}
\caption{
\textbf{(a) Passive \vs~Proactive object detection}. A learnable template encrypts the input images, which are further used to train the object detector.
\textbf{(b)} \methodName serves as a wrapper on both generic and camouflaged object detectors, enhancing the detection performance. 
\textbf{(c)} For the linear regression model under additive noise and other assumptions, the converged weights of the proactive detector are closer to the optimal weights as compared to the converged weights of the passive detector. See \cref{sec:theory} for details and proof.
%\methodName helps the object detector weights to converge at a better optimal minima, and closer to the ideal optimum weights. 
\vspace{-2mm}}
\label{fig:teaser}
\end{figure}

%All the methods mentioned above are regarded as passive methods because they take the input image as is for the task of object detection.
All these methods take input images as is for the detection task and hence are called passive methods. 
However, there is a line of research on proactive methods for a wide range of vision tasks
such as disruption~\cite{ruiz2020disrupting, segalis2020ogan}, tagging~\cite{wang2021faketagger}, manipulation detection~\cite{asnani2022proactive},  and localization~\cite{asnani2022pro_loc}.
% ~\cite{asnani2022proactive, asnani2022pro_loc, yeh2020disrupting, wang2021faketagger, segalis2020ogan}.
%proactive schemes has been adopted by recent works to perform different downstream tasks~\cite{asnani2022proactive, asnani2022pro_loc}.
Proactive methods use signals, called templates, to encrypt the input images and pass the encrypted images as the input to the network. 
% Subsequently, these encrypted images perform downstream tasks such as manipulation detection~\cite{asnani2022proactive},  manipulation localization~\cite{asnani2022pro_loc}, disruption~\cite{ruiz2020disrupting, segalis2020ogan}, and tagging~\cite{wang2021faketagger}. 
These are trained in an end-to-end manner by using either a fixed~\cite{wang2021faketagger} or learnable template~\cite{ruiz2020disrupting, segalis2020ogan,asnani2022pro_loc, asnani2022proactive} to improve the performance. 
A major advantage of proactive schemes is that such methods generalize better on unseen data/models~\cite{asnani2022proactive, asnani2022pro_loc}. Motivated by this, we propose a plug-and-play Proactive Object Detection wrapper, \methodName, to improve %the performance of 
GOD and COD detectors.  

Designing \methodName as a proactive scheme involves several challenges and key factors. 
First, the proactive wrapper needs to be a plug-and-play module that can be applied to both GOD and COD detectors. 
Secondly, the encryption process should be intuitive to benefit the object detection task. 
\forExample, an ideal template for detection should highlight the foreground objects in the input image. 
Lastly, the choice of supervision to estimate the template for encryption is hard to formulate. 
% Considering the above points, we propose a novel plug-and-play proactive wrapper, which we apply object detectors to enhance the detection performance. 

Previous proactive methods~\cite{asnani2022proactive, asnani2022pro_loc} use learnable but image-independent templates for manipulation and localization tasks.
However, the object detection task is scene-specific; therefore, the ideal template should be image-dependent. 
Based on this key insight, we propose a novel plug-and-play proactive wrapper in which we apply object detectors to enhance detection performance. 
The \methodName wrapper utilizes an encoder network to learn an image-dependent template. 
The learned template encrypts the input images by applying a transformation, defined as an element-wise multiplication between the template and the input image. 
The decoder network recovers the templates from the encrypted images. 
%We use the regression losses as supervision and use the ground-truth object map to guide the learning process to provide useful object semantics to incorporate into the template. 
We utilize regression losses for supervision and leverage the ground-truth object map to guide the learning process, thereby imparting valuable object semantics to be integrated into the template.
We then fine-tune the proactive wrapper with the GOD and COD detectors to improve their detection performance. 
Extensive experiments on MS-COCO, CAMO, \CODTenK, and \NCFourK datasets show that \methodName improves the detection performance for both GOD and COD detectors.

In summary, the contributions of this work include:
\begin{itemize}
    \item We propose a novel proactive approach \textit{\methodName} for the object detection task. To the best of our knowledge, this is the first work to develop a proactive approach to $2D$ object detection.
    %\item \methodName can be used as a plug-and-play wrapper on top of different object detectors resulting in an enhanced performance for detection/localization.
    \item We mathematically prove that the proactive method results in a better-converged model than the passive detector under assumptions and, consequently, a better object detector. 
    %\item Our experiments show that \methodName improves the performance when wrapping around different generic and camouflage object detectors.
    \item \methodName wraps around both GOD and COD detectors and improves detection performance on MS-COCO, CAMO, COD10K, and \NCFourK datasets
    %\item \methodName improves object detection performance for both GOD and COD detectors on MS-COCO, CAMO, \CODTenK, and \NCFourK datasets.
\end{itemize}

%============================================================================
%============================================================================
%============================================================================
\section{Related works}

\minisection{Proactive Schemes.} Earlier works adopt to add signals like perturbation~\cite{segalis2020ogan}, adversarial noise~\cite{ruiz2020disrupting}, and one-hot encoding~\cite{wang2021faketagger} messages while focusing on tasks like disruption~\cite{segalis2020ogan,ruiz2020disrupting } and deepfake tagging~\cite{wang2021faketagger}. Asnani~\etal~\cite{asnani2022proactive} propose to learn an optimized template for binary detection by unseen generative models. Recently, MaLP~\cite{asnani2022pro_loc} adds the learnable template to perform generalized manipulation localization for unknown generative models. Unlike these works, \methodName uses image-dependent templates and is a plug-and-play wrapper for a different task of object detection.

\minisection{Generic Object Detection}
%Previous success in face and human detection inspired researchers to tackle generic object detection. 
Detection of generic objects, instead of specific object categories such as pedestrians~\cite{pedestrian-detection-with-autoregressive-network-phases}, apples~\cite{deepapple-deep-learning-based-apple-detection-using-a-suppression-mask-r-cnn}, and others~\cite{monocular-video-based-trailer-coupler-detection-using-multiplexer-convolutional-neural-network,kumar2021groomed,kumar2022deviant}, has been a long-standing objective of computer vision.
RCNN~\cite{girshick2014rich, girshick2015region} employs the extraction of object proposals. 
He~\etal~\cite{he2015spatial} propose a spatial pooling layer to extract a fixed-length representation of all the objects. 
Modifications of RCNN~\cite{girshick2015fast, ren2015faster, li2017light, zeiler2014visualizing} increase the inference speed. 
Feature pyramid network~\cite{lin2017feature} detects objects with a wide variety of scales. 
The above methods are mostly two-stage, so inference is an issue. 
Single-stage detectors like YOLO~\cite{redmon2016you, redmon2018yolov3, bochkovskiy2020yolov4, redmon2017yolo9000,wang2022yolov7}, SSD~\cite{liu2016ssd}, HRNet~\cite{wang2020deep} and RetinaNet~\cite{lin2017focal} increase the speed and simplicity of the framework compared to the two-stage detector. 
Recently, transformer-based methods~\cite{carion2020end, zhu2020deformable} use a global-scale receptive field. Chen~\etal~\cite{chen2022diffusiondet} use diffusion models to denoise noisy boxes at every forward step. \methodName functions as a wrapper around the pre-existing object detector, facilitating its transformation into an enhanced object detector. The comparison of \methodName with prior works is summarized in \cref{tab:rel_works}. 

\minisection{Camouflaged Object Detection}
Early COD works rely on hand-crafted features like co-occurrence matrices~\cite{sengottuvelan2008performance}, $3D$ convexity~\cite{pan2011study}, optical flow~\cite{hou2011detection}, covariance matrix~\cite{ji2022fast}, and multivariate calibration components~\cite{ren2021deep}. 
Later on,~\cite{sun2021context, chen2022camouflaged} incorporate an attention-based cross-level fusion of multi-scale features to recover contextual information. 
Mei~\etal~\cite{mei2021camouflaged} take motivation by predators to identify camouflaged objects using a position and focus ideology. 
SINet~\cite{fan2020camouflaged} uses a search and identification module to perform localization. 
SINET-v2\cite{fan2021concealed} uses group-reversal attention to extract the camouflaged maps.
~\cite{kajiura2021improving} explores uncertainty maps and ~\cite{zhuge2022cubenet} utilizes cube-like architecture to integrate multi-layer features. 
ANet~\cite{le2019anabranch}, LSR~\cite{lv2021simultaneously}, and JCSOD~\cite{li2021uncertainty} employ joint learning with different tasks to improve COD. 
Lately,~\cite{mao2021transformer, yang2021uncertainty, cheng2022implicit} apply a transformer-based architecture for difficult-aware learning, uncertainty modeling, and temporal consistency. 
Zhai~\etal~\cite{zhai2021mutual} use a graph learning model to disentangle input into different features  for localization. 
DGNet~\cite{ji2023deep} uses image gradients to exploit intensity changes in the camouflaged object from the background. 
Unlike these methods, \methodName uses proactive methods to improve camouflaged object detection.

\begin{table}[t]
\centering
\caption{Comparison of \methodName with prior works.}
\label{tab:rel_works}
% \begin{adjustbox}{width=1\textwidth}
\setlength\tabcolsep{0.05cm}
\rowcolors{4}{mygray}{white}
\begin{adjustbox}{width=1\textwidth}
\begin{tabular}{l|c|c|c|c|c|c|c}
\myTopRule
\multirow{2}{*}{Method} & \multirow{2}{*}{Proactive} & \multirow{2}{*}{Task} & \multicolumn{2}{c|}{Template} & \multirow{2}{*}{COD} & \multirow{2}{*}{GOD} & \multirow{2}{*}{Plug-Play}\\
\cline{4-5}
~ & ~ & ~ & Number & Type & ~ & ~ & ~ \\
\myTopRule
Faster R-CNN~\cite{ren2015faster} & \xmark & Object Detection & - & - & \xmark & \cmark & \xmark \\
YOLO~\cite{redmon2016you} & \xmark & Object Detection & - & - & \xmark & \cmark & \xmark \\
DeTR~\cite{carion2020end} & \xmark & Object Detection & - & - & \xmark & \cmark & \xmark \\
DGNet~\cite{ji2023deep} & \xmark & Object Detection & - & - & \cmark & \xmark & \xmark \\
SINet-v2~\cite{fan2021concealed} & \xmark & Object Detection & - & - & \cmark & \xmark & \xmark \\
JCSOD~\cite{li2021uncertainty} & \xmark & Object Detection & - & - & \cmark & \xmark & \xmark \\
OGAN~\cite{segalis2020ogan} & \cmark & Disrupt & $1$ & Learnable & - & - & \xmark \\
Ruiz~\etal~\cite{ruiz2020disrupting} & \cmark & Disrupt & $1$ & Learnable & - & - & \xmark \\
Yeh~\etal~\cite{yeh2020disrupting} & \cmark & Disrupt & $1$ & Learnable & - & - & \xmark\\
FakeTagger~\cite{wang2021faketagger} & \cmark & Tagging & $\ge 1$  & Fixed, Id-dependent  & - & - & \xmark\\
Asnani~\etal~\cite{asnani2022proactive} & \cmark & Manipulation Detection & $\ge 1$ & Learnable set, Image-independent & - & - & \cmark\\
MaLP~\cite{asnani2022pro_loc}  & \cmark & Manipulation Localization & $\ge 1$  & Learnable set, Image-independent & - & - & \cmark\\
\methodName (Ours) & \cmark & Object Detection & $\ge 1$  & Learnable, Image-dependent & \cmark & \cmark & \cmark\\
\myTopRule
%\hline
\end{tabular}
\end{adjustbox}
\vspace{-2mm}
\end{table}

%============================================================================
%============================================================================
%============================================================================
\section{Proposed Approach}

Our method originates from understanding what makes proactive schemes effective. 
%Therefore, in this section, we first state the problems in \cref{sec:prob}, then present our theoretic findings in \cref{sec:theory}, and finally our method \methodName in \cref{sec:method}.
We first overview the two detection problems: GOD and COD in \cref{sec:prob}. 
We next derive \cref{lemma:1}, where we show that the proactive schemes with the multiplicative transformation of images are better than passive schemes by comparing the deviation of trained network weights from the optimal.
Based on this result, we derive that Average Precision (AP) from the proactive model is better than AP from the passive model in \cref{th:1}.
At last, we present our proactive scheme-based wrapper, \methodName, in \cref{sec:method}, which builds upon the \cref{th:1} to improve generic 2D objects and camouflaged detection.

%============================================================================
%============================================================================
\subsection{Background}\label{sec:prob}
\subsubsection{Passive Object Detection}
Although generic $2D$ object detection and camouflage detection are similar problems, they have different objective functions. Therefore, we treat them as two different problems and define their objectives separately.

\minisection{Generic 2D Object Detection.}
Let ${\vect{I}_j}$ be the set of input images given to the generic 2D object detector $\mathcal{O}$ with trainable parameters $\theta$. 
Most of these detectors output two sets of predictions per image: (1) bounding box coordinates, $\mathcal{O}(\image)_1=\hat{T} \in \mathbb{R}^4$, (2) class logits, $\mathcal{O}(\image)_2=\hat{C} \in \mathbb{R}^{C}$, where $N$ is the number of foreground object categories. If the ground-truth bounding box coordinates are $T_j$, and the ground-truth category label is $C$, the objective function of such detector is:
\begin{equation}
    \min_{\theta}\bigg\{\sum_j \Big ( ||\mathcal{O}(\image;\theta)_1-T_j||_2 \Big ) - \sum_j\sum^{N}_{i=1} \Big (C^i_j \cdot \text{log}(\mathcal{O}(\image;\theta)_2)) \Big ) \bigg\}.
    \label{eq:god_obj}
\end{equation}

\minisection{Camouflaged Object Detection.}
Let ${\vect{I}_j}$ be the input image set given to the camouflaged object detector $\mathcal{O}$ with trainable parameters $\theta$, and $\vect{G}_j$ be the ground-truth segmentation map. Prior passive works predict a segmentation map with the following objective:
\begin{equation}
    \min_{\theta}\bigg\{\sum_j \Big (\Big |\Big |\mathcal{O}(\image;\theta)-\vect{G}_j \Big |\Big |_2\Big ) \bigg\}.
    \label{eq:cod_obj}
\end{equation}

%============================================================================
%============================================================================
\subsubsection{Proactive Object Detection}
Proactive schemes~\cite{asnani2022pro_loc,asnani2022proactive} encrypt the input images with the template to aid manipulation detection/localization. 
Such schemes take an input image $\image\in \mathbb{R}^{H\times W\times 3}$ and learns a 
template $\vect{S}_j\in \mathbb{R}^{H\times W}$.
\methodName uses image-dependent templates to improve object detection.
Given an input image $\image\in \mathbb{R}^{H\times W\times 3}$, \methodName learns to output a template $\vect{S}_j\in \mathbb{R}^{H\times W}$, which can be used by a transformation $\mathcal{T}$ resulting in encrypted images $\mathcal{T}(\image)$. 
\methodName uses element-wise multiplication as the transformation $\mathcal{T}$, which is defined as: 
\begin{equation}
    \mathcal{T}(\image) = \mathcal{T}(\image;\vect{S}_j)={\image}\odot{\vect{S}_j}.
    \label{eq:trans}
\end{equation}

%============================================================================
%============================================================================
\subsection{Mathematical Analysis of Passive and Proactive Detectors}\label{sec:theory}
\methodName optimizes the template to improve the performance of the object detector. 
We argue that this template helps arrive at a better global minima representing the optimal parameters $\theta$. 
We now define the following lemma to support our argument:

\begin{lemma}\label{lemma:1}
\textbf{Converged weights of proactive and passive detectors.}
Consider a linear regression model that regresses an input image $\image$ under an additive noise setup to obtain the 2D coordinates.
Assume the noise under consideration $e$ is a normal random variable $\mathcal{N}(0,\sigma^2)$.
Let $\vect{w}$ and $\vect{w}^{*}$ denote the trained weights of the pretrained linear regression model and the optimal weights of the linear regression model. 
% Assume the error in bounding box prediction $e$ is a normal random variable $\mathcal{N}(0,\sigma^2)$, and a SGD optimizer is used to update the model parameters with an $L_1$ loss. 
Also, assume SGD optimizes the model parameters with decreasing step size $\step$ such that the steps are square summable \thatIs, $\stepSum\!=\!\lim\limits_{\instant \rightarrow \infty} \sum\limits_{k=1}^\instant \step_k^2$ exist, and the noise is independent of the image.
Then, there exists a template $\vect{S}_j\in[0,1]$ for the image $\image$ such that the multiplicative transformation of images as the input results in a trained weight $\vect{w}'$ closer to the optimal weight than the originally trained weight $\vect{w}$.
% which results in a better set of weights.
%which when used to encrypt the images by applying a transformation $\mathcal{T}(\image;\vect{S}_j)={\image} \times {\vect{S}_j}$, would result in a better global minima of optimal weights $\vect{w}'$. Therefore, we would have:
In other words,
\begin{equation}
    \mathbb{E}(||\vect{w}'-\vect{w}^{*}||_2)<\mathbb{E}(||\vect{w}-\vect{w}^{*}||_2).
\end{equation}
\end{lemma}
% \begin{proof}
    The proof of \cref{lemma:1} is in supplementary.
    We use the variance of the gradient of the encrypted images to arrive at this lemma.
% \end{proof}
We next use \cref{lemma:1} to derive the following theorem:

\begin{theorem}\label{th:1}
\textbf{AP comparison of proactive and passive detectors.}
Consider a linear regression model that regresses an input image $\image$ under an additive noise setup to obtain the 2D coordinates.
Assume the noise under consideration $e$ is a normal random variable $\mathcal{N}(0,\sigma^2)$.
Let $\vect{w}$ and $\vect{w}^{*}$ denote the trained weights of the pretrained linear regression model and the optimal weights of the linear regression model. 
Also, assume SGD optimizes the model parameters with decreasing step size $\step$ such that the steps are square summable \thatIs, $\stepSum\!=\!\lim\limits_{\instant \rightarrow \infty} \sum\limits_{k=1}^\instant \step_k^2$ exist, and the noise is independent of the image.
Then, the AP of the proactive detector is better than the AP of the passive detector.
% Given an input image $\image$, let  the optimal weights of the pretrained linear regression model $\vect{w}$, the optimal weights of the ideal linear regression model $\vect{w}^{*}$. 
% Assume the error in bounding box prediction $e$ is a normal random variable $\mathcal{N}(0,\sigma^2)$, and a SGD optimizer is used to update the model parameters with an $L_1$ loss. Assume that an optimal template $\vect{S}_j$ bounded by $[0,1]$, is used to encrypt the images by applying a transformation $\mathcal{T}(\image;\vect{S}_j)={\image} \times {\vect{S}_j}$. Assume the optimal weights of the linear regression model trained with input images $\mathcal{T}(\image;\vect{S}_j)$ be $\vect{w}'$. Then the mean average precision (mAP) would be worse for pretrained model with weights $\vect{w}$ compared to fine-tuned model with weights $\vect{w}'$.
\end{theorem}
% \begin{proof}
    The proof of \cref{th:1} is in the supplementary.
    We use the \cref{lemma:1} and the non-decreasing nature of AP~\wrt~\iou to arrive at this theorem.
% \end{proof}
% Guided by \cref{th:1} proactive schemes, 
Next, we adapt the objectives of \cref{eq:god_obj,eq:cod_obj} to incorporate the proactive methods as follows: 
\begin{equation}
    \min_{\theta, \vect{S}_j}\bigg\{\sum_j \Big ( ||\mathcal{O}(\mathcal{T}(\image;\vect{S}_j);\theta)_1-T_j||_2 \Big ) - \sum_j\sum^{N}_{i=1} \Big (C^i_j \cdot \text{log}(\mathcal{O}(\mathcal{T}(\image;\vect{S}_j);\theta)_2) \Big ) \bigg\},
    \label{eq:god_obj_pro}
\end{equation}
\begin{equation}
    \min_{\theta, \vect{S}_j}\bigg\{\sum_j \Big (\Big |\Big |\mathcal{O}(\mathcal{T}(\image;\vect{S}_j);\theta)-\vect{G}_j \Big |\Big |_2\Big ) \bigg\}.
    \label{eq:cod_obj_pro}
\end{equation}

\begin{figure}[t!]
\centering
\includegraphics[trim={0 -4 0 0},clip,width=1\textwidth]{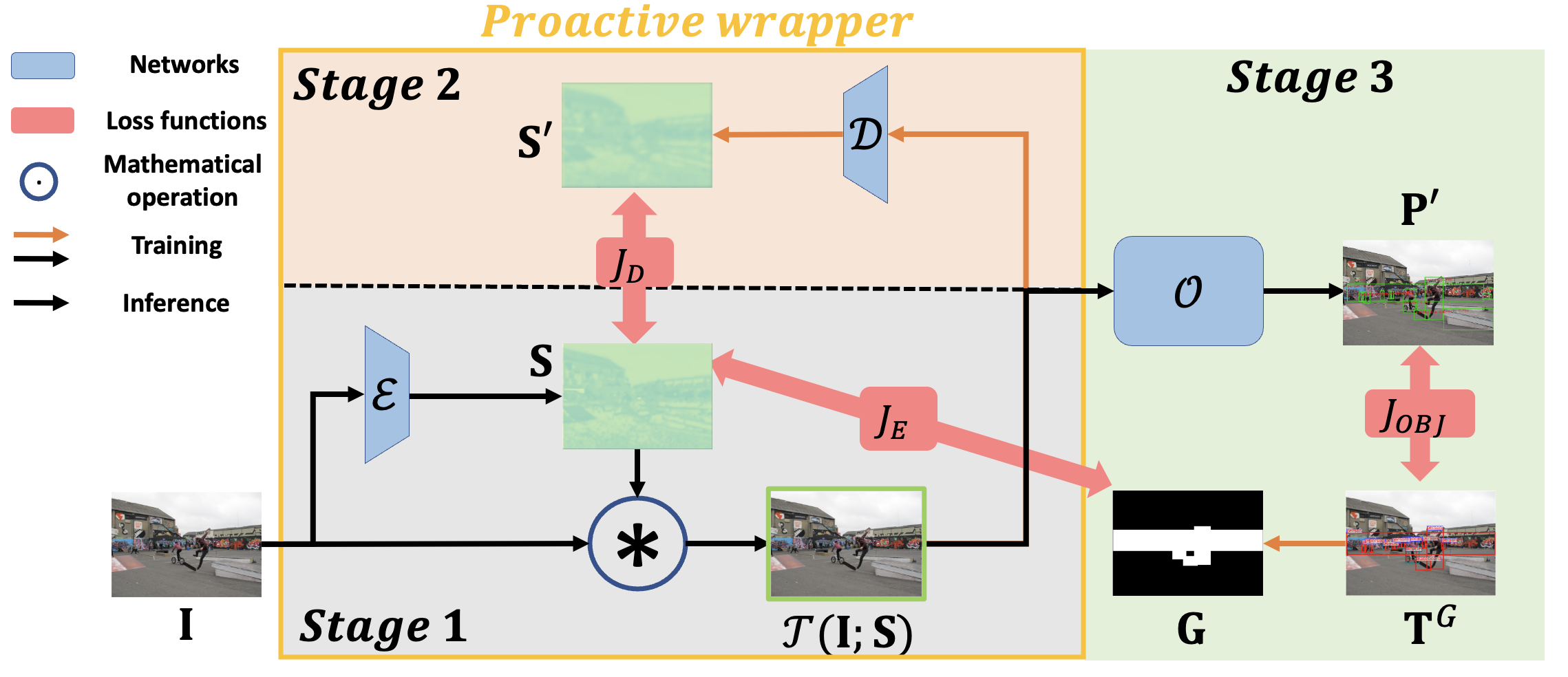}
\caption{\textbf{Overview of \methodName.} \methodName consists of three stages: (1) template generation, (2) template recovery, and (3) detector fine-tuning. The templates are generated by encoder network $\mathcal{E}$ to encrypt the input images. The decoder network $\mathcal{D}$ is used to recover the template from the encrypted images. Finally, the encrypted images are used to fine-tune the object detector to perform detection. We train all the stages in an end-to-end manner. However, for inference, we only use stages $1$ and $3$. Best viewed in color.}
\label{fig:overview}
\vspace{-2mm}
\end{figure}

%============================================================================
%============================================================================
%============================================================================
\subsection{\methodName}\label{sec:method}

Our proposed approach comprises of three stages: template generation, template recovery, and detector fine-tuning. First, we use an encoder network to generate an image-dependent template for image encryption. 
This encrypted image is further used to recover the template through a decoder network. Finally, the object detector is fine-tuned using the encrypted images. All three stages are trained in an end-to-end fashion. While all the stages are used for training \methodName, we specifically use only stages $1$ and $3$ for inference. We will now describe each stage in detail. 

\subsubsection{Proactive Wrapper}
Our proposed approach consists of three stages, as shown in Fig.~\ref{fig:overview}. However, only the first two stages are part of our proposed proactive wrapper, which can be applied to object detector to improve its performance.

\minisection{Stage 1: Template Generation.}
Prior works learn a set of templates~\cite{asnani2022pro_loc, asnani2022proactive} in their proactive schemes. 
This set of templates is enough to perform the respective downstream tasks as the generative model manipulates the template, which is easy to capture with a set of learnable templates. 
%However, for object detection task, every image has different semantics of the objects which varies a lot with size, appearance, color, etc. 
However, for object detection tasks, every image has unique object characteristics such as size, appearance, and color that can vary significantly.
This variability present in the images may exceed the descriptive capacity of a finite set of templates, thereby necessitating the use of image-specific templates to accurately represent the range of object features present in each image. 
In other words, a fixed set of templates may not be sufficiently flexible to capture the diversity of visual features across the given set of input images, thus demanding more adaptable, image-dependent templates.

Motivated by the above argument, we propose to generate the template $\vect{S}_j$ for every image using an encoder network. We hypothesize that highlighting the area of the key foreground objects would be beneficial for object detection. Therefore, for GOD, we use the ground-truth bounding boxes $T^G$ to generate the pseudo ground-truth segmentation map. Specifically, for any image $\image$, if the bounding box coordinates are $T^G_j=\{x_1, x_2, y_1, y_2\}$, we define the pseudo ground-truth segmentation map as:
\begin{align*}
&\forall m\in [0,H],n\in[0,W] \text{, we have}\\
\vect{G}_j(m,n)=1&\text{ if }x_1\le m \le x_2\text{ and }y_1\le n \le y_2, \text{ otherwise } 0
\end{align*}

However, for COD, the dataset already has the ground-truth segmentation map $\vect{G}_j$, which we use as the supervision for the encoder to output the templates with semantic information of the image to be restricted only in the region of interest for the detector. For both GOD and COD, we minimize the cosine similarity (Cos) between $\vect{S}_j$ and $\vect{G}_j$ as the supervision for the encoder network. The encoder loss $J_E$ is as follows:
\begin{equation}
J_E = 1-\text{Cos}(\vect{S}_j,\vect{G}_j)=1-\text{Cos}(\mathcal{E}(\image),\vect{G}_j).
\label{eq:L_JE}  
\end{equation}

This generated template acts as a mask for the input image to highlight the object region of interest for the detector. We use this template with the transformation $\mathcal{T}$ to encrypt the input image as $\mathcal{T}(\image;\vect{S}_j)={\image} \odot {\vect{S}_j}$. As we start from the pretrained model of object detector $\mathcal{O}$, we initialize the bias of the last layer of the encoder as $0$ so that for the first few iterations, $\vect{S}_j\approx \vect{1}$. This is to ensure that the distribution of $\image$ and $\mathcal{T}(\image;\vect{S}_j)$ remains similar for the first few iterations, and $\mathcal{O}$ doesn't encounter a sudden change in its input distribution.

\minisection{Stage 2: Template Recovery.}
So far, we have discussed the generation of template $\vect{S}_j$ using $\mathcal{E}$, which will be used as a mask to encrypt the input image. The encrypted images are used for two purposes: (1) recovery of templates and (2) fine-tuning of the object detector. 
The main intuition of recovering the templates is from the prior works on image steganalysis~\cite{ur2019end, rahim2018end} and proactive schemes~\cite{asnani2022pro_loc, asnani2022proactive}. Motivated by these works, we draw the following insight: 
\textit{``To properly learn the optimal template and embed it onto the input images, it is beneficial to recover the template from encrypted images."} 

To perform recovery, we exploit an encoder-decoder approach. Using this approach leverages the strengths of the encoder network $\mathcal{E}$ for feature extraction, capturing the most useful salient details, and the decoder network $\mathcal{D}$ for information recovery, allowing for efficient and effective encryption and decryption of the template. We also empirically show that not using the decoder to recover the templates harms the  object detection performance.

To supervise $\mathcal{D}$ in recovering $\vect{S}_j$ from $\mathcal{T}(\image;\vect{S}_j)$, we propose to maximize the cosine similarity between the recovered template, $\vect{S}^{'}_j$ and $\vect{S}_j$. The decoder loss is as follows:
\begin{equation}
J_D = 1-\text{Cos}(\vect{S}^{'}_j, \vect{S}_j)=1-\text{Cos}(\mathcal{D}(\mathcal{T}(\image;\vect{S}_j)), \vect{S}_j).
\label{eq:L_JD}  
\end{equation}

\minisection{Stage 3: Detector Fine-tuning.}
Due to our encryption, the distribution of the images input to the pretrained $\mathcal{O}$ changes. Thus, we fine-tune $\mathcal{O}$ on the encrypted images $\mathcal{T}(\image;\vect{S})$. As proposed in Theorem $1$, given the encrypted images $\mathcal{T}(\image;\vect{S})$, we use the pretrained detector $\mathcal{O}$ with parameters $\theta$ to arrive at a better local minima. Therefore, the general objective of GOD and COD in \cref{eq:god_obj_pro} and \cref{eq:cod_obj_pro} change to as follows:
\begin{equation}
\footnotesize
\min_{\theta, \theta_{\mathcal{E}}, \theta_{\mathcal{D}}}\bigg\{\sum_j \Big ( ||\mathcal{O}(\mathcal{T}(\image;\mathcal{E}(\image;\theta_{\mathcal{E}}));\theta,\theta_{\mathcal{D}})_1-T_j||_2 - \sum^{N}_{i=1} \big (C^i_j. \text{log}(\mathcal{O}(\mathcal{T}(\image;\mathcal{E}(\image;\theta_{\mathcal{E}}));\theta,\theta_{\mathcal{D}})_2) \big )\Big ) \bigg\},
    \label{eq:god_obj_pro_fin}
\end{equation}
\begin{equation} 
\footnotesize
    \min_{\theta, \theta_{\mathcal{E}}, \theta_{\mathcal{D}}}\bigg\{\sum_j \Big (\Big |\Big |\mathcal{O}(\mathcal{T}(\image;\mathcal{E}(\image;\theta_{\mathcal{E}}));\theta,\theta_{\mathcal{D}})-\vect{G}_j \Big |\Big |_2\Big ) \bigg\}.
    \label{eq:cod_obj_pro_fin}
\end{equation}

We use the detector-specific loss function $J_{OBJ}$ of $\mathcal{O}$ along with the encoder and decoder loss in \cref{eq:L_JE} and \cref{eq:L_JD} to train all the three stages. The overall loss function $J$ to train \methodName is as follows:
\begin{align}
J &=\lambda_{OBJ}J_{OBJ}+\lambda_EJ_E+\lambda_D{J_D}.
\label{eqn:ovr_loss}
\end{align}

\begin{table}[t!]
% \small
%\begin{center}
\caption{\textbf{GOD results} on MS-COCO val split. \methodName improves the performance of all GOD at all thresholds and across all categories.} 
\label{tab:baseline_comp_proac_god}
\centering
 \begin{adjustbox}{width=0.93\textwidth}
\setlength{\tabcolsep}{0.12cm}
\rowcolors{3}{mygray}{white}
\begin{tabular}{l|c c c|c c c}
\myTopRule
%\hline
Method & AP$~\bf\uparrow$ & AP$_{50}\bf\uparrow$ &  AP$_{75}\bf\uparrow$ & AP$_S\bf\uparrow$ & AP$_M\bf \uparrow$& AP$_L\bf\uparrow$ \\ 
\myTopRule
Faster R-CNN~\cite{ren2015faster} & $19.3$ & $42.5$ & $16.9$ & $1.8$ & $17.9$ & $39.3$\\
Faster R-CNN~\cite{ren2015faster}$+$\methodName & $\bf31.7$ & $\bf52.6$ &  $\bf 33.3$ & $\bf11.0$ & $\bf35.5$ & $\bf51.1$\\ 
\hline\myTopRule
Faster R-CNN $+$ FPN~\cite{lin2017feature} & $37.3$ & $58.0$ & $40.6$ & $21.4$ & $41.0$ & $48.4$\\
Faster R-CNN $+$ FPN~\cite{lin2017feature} $+$ Seg. Mask~\cite{he2017mask} & $38.2$ & $60.3$ & $41.7$ & $22.1$ & $43.2$ & $\bf51.2$\\
Faster R-CNN $+$ FPN~\cite{lin2017feature} $+$ \methodName & $\bf38.5$ & $\bf60.4$ &  $\bf 41.9$ & $\bf22.5$ & $\bf43.4$ & $49.8$\\ 
\hline\myTopRule
Sparse R-CNN~\cite{sun2021sparse}& $37.6$ & $55.6$  & $40.2$ & $20.5$ & $39.6$ & $52.9$  \\
Sparse R-CNN~\cite{sun2021sparse}$+$ \methodName & $\bf39.2$ & $\bf57.5$ &  $\bf41.5$ & $\bf21.7$ & $\bf40.1$ & $\bf53.6$\\\hline\myTopRule
YOLOv5~\cite{redmon2016you} & $48.9$ & $67.6$  & $53.1$ & $31.8$ & $54.4$ & $62.3$  \\
YOLOv5~\cite{redmon2016you}$+$ \methodName & $\bf49.4$ & $\bf67.9$ &  $\bf53.5$ & $\bf32.0$ & $\bf55.1$ & $\bf62.6$\\
\hline\myTopRule
DeTR~\cite{carion2020end} & $41.9$ & $62.3$  & $44.1$ & $20.3$ & $45.8$ & $61.0$\\
DeTR~\cite{carion2020end}$+$ \methodName & $\bf42.1$ & 
 $\bf62.6$ & $\bf44.4$ & $\bf20.4$ & $\bf46.0$ & $\bf61.3$ \\
\myTopRule
\end{tabular}
\end{adjustbox}
%\end{center}
\vspace{-2mm}
\end{table}

\begin{table}[t!]
% \small
%\begin{center}
\caption{\textbf{COD results} on CAMO, \CODTenK and \NCFourK datasets. \methodName outperforms DGNet on all datasets and metrics.} 
\label{tab:baseline_comp_proac_cod}
\centering
\begin{adjustbox}{width=0.95\textwidth}
\setlength{\tabcolsep}{0.05cm}
\rowcolors{4}{mygray}{white}
\begin{tabular}{l|c c c c| c c c c|c c c c}
\myTopRule
%\hline
%\rowcolor{mygray}
& \multicolumn{4}{c|}{CAMO} & \multicolumn{4}{c|}{\CODTenK} & \multicolumn{4}{c}{\NCFourK}\\ \hhline{~|-|-|-|-|-|-|-|-|-|-|-|-}
%\rowcolor{mygray}
\multirow{-2}{*}{Method} & E$_m\!\uparrow$ & S$_m\!\uparrow$ & wF$_{\beta}\!\uparrow$ & MAE$\downarrow$ & E$_m\!\uparrow$ & S$_m\!\uparrow$ & wF$_{\beta}\!\uparrow$ & MAE$\downarrow$ & E$_m\!\uparrow$ & S$_m\!\uparrow$ & wF$_{\beta}\!\uparrow$ & MAE$\downarrow$\\
\myTopRule
DGNet\cite{ji2023deep} & $0.859$ & $0.791$ & $0.681$ & $0.079$ & $0.833$ & $0.776$ & $0.603$ & $0.046$ & $0.876$ & $0.815$ & $0.710$ & $0.059$\\
$+$ \methodName & $\bf0.871$ & $\bf0.797$ & $\bf0.702$ & $\bf0.071$ & $\bf 0.869$ & $\bf0.803$ & $\bf0.661$ & $\bf0.037$ & $\bf 0.900$ & $\bf0.838$ & $\bf0.755$ & $\bf0.049$\\
\myTopRule
%\hline
%\hline
\end{tabular}
\end{adjustbox}
%\end{center}
\vspace{-2mm}
\end{table}

%============================================================================
%============================================================================
%============================================================================
\section{Experiments}
We apply \methodName for two categories of object detectors: GOD and COD. 

\minisection{GOD Baselines.}For GOD, we apply \methodName on four detectors with varied architectures: two-stage, one-stage, and transformer-based detectors, namely, Faster R-CNN~\cite{ren2015faster}, YOLO~\cite{redmon2016you}, Sparse R-CNN, and DeTR~\cite{carion2020end}.
We use these works as baselines for three reasons: (1) varied architecture types, (2) their increased prevalence in the community, and (3) varied timelines (from earlier to recent detectors). 
We use the PyTorch~\cite{paszke2019pytorch} code of the respective detectors for our GOD experiments and use the corresponding GODs as our baseline. 
For YOLOv5 and DeTR, we use the official repositories released by the authors; for Faster R-CNN, we use the public repository "Faster R-CNN.pytorch". For other GOD detectors, we use Detectron2 library as the pre-trained detector. We use the ResNet101 backbone for Faster R-CNN, Sparse R-CNN and DeTR, and CSPDarknet53 for YOLOv5. 

\minisection{COD Baselines.}For COD, we apply \methodName on the current SoTA camouflage detector DGNet~\cite{ji2023deep} and use DGNet as our baseline. 
For all object detectors, we use the pretrained model released by the authors and fine-tune them with \methodName. Please see the supplementary for more details.

\minisection{Datasets.}
% We use a number of datasets in our training and inference. 
Our experiments use the MS-COCO $2017$~\cite{lin2014microsoft} dataset for GOD, while we use CAMO~\cite{le2019anabranch}, \CODTenK~\cite{fan2021concealed }, and \NCFourK~\cite{lv2021simultaneously} datasets for COD. 
We use the following splits of these datasets:
\begin{itemize}
    \item MS-COCO $2017$ Val Split~\cite{lin2014microsoft}: It includes $118{,}287$ images for training and $5K$  for testing.
    \item \CODTenK Val Split~\cite{fan2021concealed }: It includes $4{,}046$ camouflaged images for training and $2{,}026$  for testing. 
    \item CAMO Val Split~\cite{le2019anabranch}: It includes $1K$ camouflaged images for training and $250$ for testing. 
    \item \NCFourK Val~\cite{lv2021simultaneously}: It includes $4{,}121$ \NCFourK images. We use it for generalization testing as in~\cite{ji2023deep}. 
\end{itemize}

%============================================================================
%============================================================================

\minisection{Evaluation Metrics.} We use mean average precision average at multiple thresholds in $[0.5,0.95]$ (AP) for GOD as in \cite{lin2014microsoft}. We also report results at threshold of $0.5$ (AP$_{50}$), threshold of $0.75$ (AP$_{75}$) and at different object sizes: small (AP$_{S}$), medium (AP$_{M}$), and large (AP$_{L}$). For COD, we use E-measure $E_m$, S-measure $S_m$, weighted F1 score $wF_{\beta}$ and mean absolute error $MAE$ as~\cite{ji2023deep}.  

\subsection{GOD Results}
\minisection{Quantitative Results.}\cref{tab:baseline_comp_proac_god} shows the results of applying \methodName on GOD networks. 
\methodName improves the average precision of all three detectors. The performance gain is significant for Faster R-CNN. 
As Faster R-CNN is an older detector, it was at a worse minima to start with. \methodName improves the convergence weight of Faster R-CNN by a significant margin, thereby improving the performance. 
We further experiment with two variations of Faster R-CNN, namely, Faster R-CNN + FPN and Sparse-RCNN. We observe an increase in the performance of both detectors. 
\methodName also improves newer detectors like YOLOv5 and DeTR, although the gains are smaller compared to Faster R-CNN. 
We believe this happens because the newer detectors leave little room for improvement due to which \methodName improves the performance slightly. We next compare \methodName with a work that leverage segmentation map as a mask for object detection. We compare our performance with Mask R-CNN~\cite{he2017mask}, which uses an image segmentation branch to help with object detection. \cref{tab:baseline_comp_proac_god} shows that the gains using Mask R-CNN are lower than using our proactive wrapper.

\begin{figure}[t!]
\centering
\includegraphics[trim={0 -4 0 0},clip,width=1\textwidth]{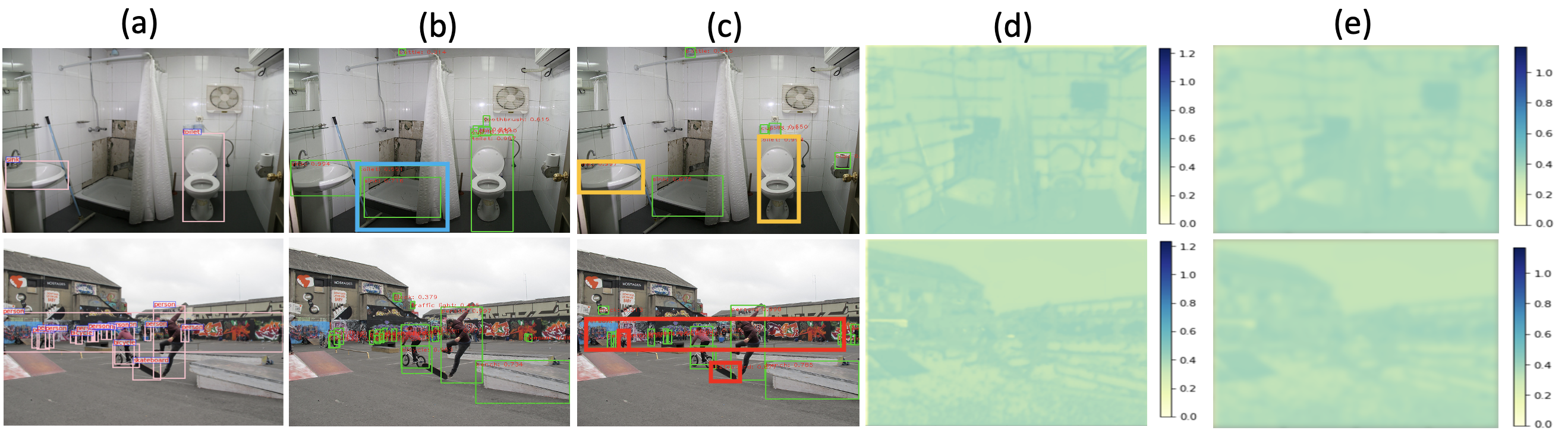}
\caption{\textbf{Qualitative GOD Results} on MS-COCO $2017$ dataset. 
%Visualization of Faster R-CNN predictions on MS-COCO $2017$ after applying \methodName. 
(a) ground-truth annotations, (b) Faster R-CNN \cite{ren2015faster} predictions, (c) Faster R-CNN \cite{ren2015faster}$+$ \methodName predictions, (d) generated template, and (e) recovered template. We highlight the objects responsible for improvement in (c) as compared to (b). The yellow box represents better localization, the blue box represents false positives, and the red box represents missed predictions. \methodName improves on all these errors made by (b). }
\label{fig:god_visual}
\vspace{-3mm}
\end{figure}

\minisection{Qualitative Results.}\cref{fig:god_visual} shows qualitative results for the MS-COCO $2017$ dataset. \methodName clearly improves the performance of pretrained Faster R-CNN for three types of errors: Missed predictions, false negatives, and localization errors. \methodName has a lower number of missed predictions, fewer false positives, and better bounding box localization. 
We also visualize the generated and recovered templates. We see that the template has object semantics of the input images. When the template is multiplied with the input image, it highlights the foreground objects, thereby making the task of object detector easier. 

\minisection{Error Analysis.}We show the error analysis~\cite{bolya2020tide} for GOD section $4$ of the supplementary. We observe that all GOD detectors make mistakes mainly due to five types of errors: classification, localization, duplicate detection, background detection, and missed detection. The main reason for the degraded performance is the errors in which the foreground-background boundary is missed. These errors include localization, background detection, and missed detection. Our proactive wrapper significantly corrects these errors, as the template has object semantics, which, when multiplied with the input image, highlights the foreground objects, consequently simplifying the task of object detection.

\begin{figure}[t!]
\centering
\includegraphics[trim={0 -4 0 0},clip,width=1\textwidth]{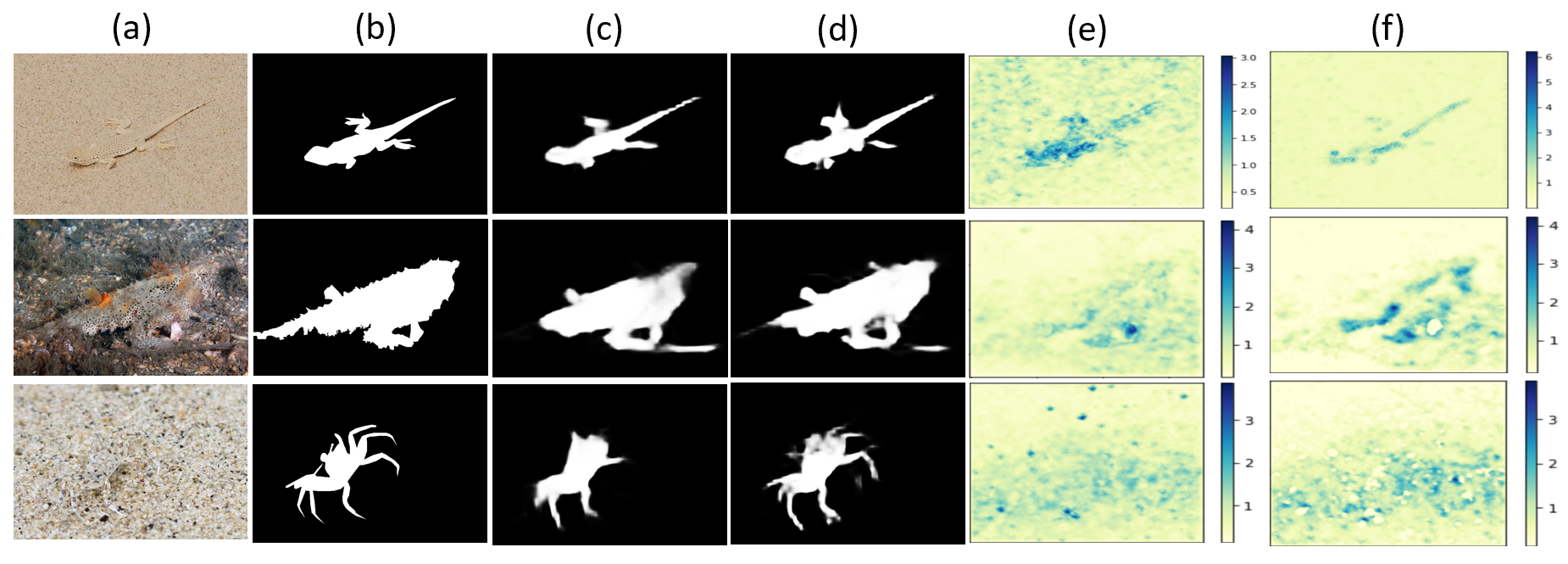}
\caption{\textbf{Qualitative COD Results} 
%Visualization of images in 
on CAMO, \CODTenK, and \NCFourK datasets from top to bottom, after applying \methodName. (a) input images, (b) ground-truth camouflaged map, (c) DGNet\cite{ji2023deep} predictions, (d) DGNet\cite{ji2023deep}$+$ \methodName predictions, (e) generated \methodName template, and (f) recovered \methodName template. \methodName template has the semantics of the camouflaged object, which aids DGNet in detection.}
\label{fig:cod_visual}
\vspace{-2mm}
\end{figure}

\iffalse
\begin{table}[t]
% \small
\caption{ Generalization performance with prior camouflaged object detection works. } 
\label{tab:gen_cod}
\centering
\begin{adjustbox}{width=1\textwidth}
\begin{tabular}{c|c|c c c c|c c c c|c c c c}
\hline\hline
\rowcolor{mygray}  &  & \multicolumn{4}{c|}{CAMO} & \multicolumn{4}{c|}{\CODTenK} & \multicolumn{4}{c}{\NCFourK}\\ \hhline{~|~|-|-|-|-|-|-|-|-|-|-|-|-}
\rowcolor{mygray} \multirow{-2}{*}{Method}& \multirow{-2}{*}{Training data} & $E_m\bf \uparrow$ & $S_m\bf \uparrow$ & $wF_{\beta}\bf \uparrow$ & $MAE\bf \downarrow$ & $E_m\bf \uparrow$ & $S_m\bf \uparrow$ & $wF_{\beta}\bf \uparrow$ & $MAE\bf \downarrow$& $E_m\bf \uparrow$ & $S_m\bf \uparrow$ & $wF_{\beta}\bf \uparrow$ & $MAE\bf \downarrow$\\\hline
 DGNet\cite{ji2023deep} &  & & & & & & & & & & & & \\
$+$ \methodName & \multirow{-2}{*}{CAMO} & & & & & & & & & & & &\\
\hline
\rowcolor{mygray} DGNet~\cite{ji2023deep} &  & & & & & & & & & & & & \\
\rowcolor{mygray} $+$ \methodName & \multirow{-2}{*}{\CODTenK} & & & & & & & & & & & &\\
\hline
DGNet~\cite{ji2023deep} & \multirow{2}{*}{CAMO $+$ \CODTenK} & & & & & & & & & & & &\\
$+$ \methodName & & & & & & & & & & & & &\\\hline
\hline
%\hline
\end{tabular}
\end{adjustbox}
\vspace{-2mm}
\end{table}
\fi

\begin{table}[t]
% \small
%\begin{center}
\caption{Performance comparison with proactive works. MaLP~\cite{asnani2022pro_loc} has a significantly deteriorated performance than \methodName.}
\label{tab:baseline_comp_proac}
\centering
\begin{adjustbox}{width=0.95\textwidth}
\setlength{\tabcolsep}{0.05cm}
\rowcolors{4}{mygray}{white}
\begin{tabular}{l|c c c c| c c c c|c c c c}
\myTopRule
%\hline
%\rowcolor{mygray}
& \multicolumn{4}{c|}{CAMO} & \multicolumn{4}{c|}{\CODTenK} & \multicolumn{4}{c}{\NCFourK}\\ \hhline{~|-|-|-|-|-|-|-|-|-|-|-|-}
%\rowcolor{mygray}
\multirow{-2}{*}{Method} & E$_m\!\uparrow$ & S$_m\!\uparrow$ & wF$_{\beta}\!\uparrow$ & MAE$\downarrow$ & E$_m\!\uparrow$ & S$_m\!\uparrow$ & wF$_{\beta}\!\uparrow$ & MAE$\downarrow$ & E$_m\!\uparrow$ & S$_m\!\uparrow$ & wF$_{\beta}\!\uparrow$ & MAE$\downarrow$\\
\myTopRule
MaLP~\cite{asnani2022pro_loc} & $0.474$ & $0.514$ & $0.218$ & $0.254$  & $0.491$ & $0.520$ & $0.150$ & $0.202$ & $0.503$ & $0.548$ & $0.228$ & $0.222$\\
\methodName & $\bf0.871$ & $\bf0.797$ & $\bf0.702$ & $\bf0.071$ & $\bf 0.869$ & $\bf0.803$ & $\bf0.661$ & $\bf0.037$ & $\bf 0.900$ & $\bf0.838$ & $\bf0.755$ & $\bf0.049$\\
\myTopRule
%\hline
%\hline
\end{tabular}
\end{adjustbox}
%\end{center}
\vspace{-2mm}
\end{table}

\begin{table}[t]
% \small
%\begin{center}
\centering
\caption{\textbf{Ablation studies} of \methodName using Faster R-CNN GOD on MS-COCO $2017$ dataset. 
Removing the encoder/decoder network or adding the template results in degraded performance.} 
\label{tab:abl_framework}
\begin{adjustbox}{width=0.95\textwidth}
\setlength{\tabcolsep}{0.14cm}
% \rowcolors{3}{mygray}{white}
\begin{tabular}{l|l|c c c|c c c}
\myTopRule
Changed & From$\rightarrowRHD$To  & AP$~\bf\uparrow$ &AP$_{50}\bf\uparrow$ &  AP$_{75}\bf\uparrow$ & AP$_S\bf\uparrow$ & AP$_M\bf \uparrow$& AP$_L\bf\uparrow$ \\ 
\myTopRule
\multirow{2}{*}{Template} & Image Dependent$\rightarrowRHD$Fixed & $17.6$ & $37.9$ & $15.1$ & $1.3$ & $15.4$ & $39.5$\\
& Image Dependent$\rightarrowRHD$Universal & $19.4$ & $42.6$ &  $17.1$ & $1.9$ & $18.0$ & $39.4$\\
\hline
Decoder & Yes$\rightarrowRHD$No & $25.2$ & $46.1$ &  $26.2$ & $5.3$ & $26.6$ & $24.1$\\
\hline
Transformation & Multiply$\rightarrowRHD$Add & $19.2$ & $42.3$ &  $20.1$ & $1.7$ & $17.9$ & $39.1$\\
\hline
\rowcolor{mygray} \methodName & - &$\bf31.7$ & $\bf52.6$ &  $\bf33.3$ & $\bf11.0$ & $\bf35.5$ & $\bf51.1$\\
\myTopRule
% \hline
% \hline
\end{tabular}
\end{adjustbox}
%\end{center}
\vspace{-1mm}
\end{table}

\begin{table}[t]
%\rowcolors{1}{mygray}{white}
%\small
%\begin{center}
\centering
\caption{\textbf{Ablation of training iterations} on Faster R-CNN. YOLOv5, and DeTR for more iterations similar to after applying \methodName. We also report the inference time for all the detectors before and after applying \methodName. 
%Training the object detector for more iterations does help, but only upto a certain extent. 
Training object detectors proactively with \methodName results in more performance gain compared to training passively for more iterations. \methodName adds an overhead cost on top of the inference cost of detectors. } 
\label{tab:god_more_train}
\centering
\begin{adjustbox}{width=1\textwidth}
\setlength{\tabcolsep}{0.2cm}
% \rowcolors{2}{mygray}{white}
\begin{tabular}{l|c|c c c| c c c| c}
\myTopRule
Method & Iterations & AP$~\bf\uparrow$ & AP$_{50}\bf\uparrow$  & AP$_{75}\bf\uparrow$ & AP$_S\bf\uparrow$ & AP$_M\bf \uparrow$& AP$_L\bf\uparrow$ & Time ($ms$)\\  
\myTopRule
Faster R-CNN~\cite{ren2015faster} & $1\times$ & $19.3$ & $42.5$ &  $16.9$ & $1.8$ & $17.9$ & $39.3$ & \multirow{2}{*}{$161.1$}\\
Faster R-CNN~\cite{ren2015faster} & $2\times$ & $20.1$ & $46.6$ &  $21.5$ & $3.3$ & $20.3$ & $41.2$ &\\ 
\rowcolor{mygray}Faster R-CNN~\cite{ren2015faster} $+$ \methodName & $2\times$ & $\bf31.7$ & $\bf52.6$ &  $\bf33.3$ & $\bf11.0$ & $\bf35.5$ & $\bf51.1$ & $175.3$ $ (\uparrow8.7\%)$\\\hline
YOLOv5~\cite{redmon2016you} & $1\times$ &$48.9$ & $67.6$ &  $53.1$ & $31.8$ & $54.4$ & $62.3$ & \multirow{2}{*}{$48.5$}\\
YOLOv5~\cite{redmon2016you} & $2\times$ &$48.8$ & $67.7$ &  $53.0$ & $31.8$ & $54.7$ & $62.4$ &\\ 
\rowcolor{mygray}YOLOv5~\cite{redmon2016you} $+$ \methodName & $2\times$ & $\bf49.4$ & $\bf67.9$ &  $\bf53.5$ & $\bf32.0$ & $\bf55.1$ & $\bf62.6$ & $62.7$ $ (\uparrow29.1\%)$\\\hline
DeTR~\cite{carion2020end} & $1\times$ &$41.9$ & $62.3$ &  $44.1$ & $20.3$ & $45.8$ & $61.0$ & \multirow{2}{*}{$194.2$}\\
DeTR~\cite{carion2020end} & $2\times$ & $41.9$ & $62.4$ & $44.0$ & $20.1$ & $45.9$ & $61.1$ & \\ 
\rowcolor{mygray}DeTR~\cite{carion2020end} $+$ \methodName & $2\times$ & $\bf42.1$ & $\bf62.6$  & $\bf44.4$ & $\bf20.4$ & $\bf46.0$ & $\bf61.3$ & $208.4$ $ (\uparrow7.2\%)$\\
\myTopRule
%\hline
%\hline
\end{tabular}
\end{adjustbox}
%\end{center}
\vspace{-1mm}
\end{table}

%============================================================================
%============================================================================
\subsection{COD Results}
\minisection{Quantitative Results.} \cref{tab:baseline_comp_proac_cod} shows the result of applying \methodName to DGNet~\cite{ji2023deep} on three different datasets. \methodName, when applied on top of DGNet, outperforms DGNet on all four metrics for all datasets. 
The biggest gain appears in \CODTenK and \NCFourK datasets.
This is impressive as these datasets have more diverse testing images than CAMO. 
As \NCFourK is only a testing set, the higher performance of \methodName demonstrates its superior generalizability as compared to DGNet~\cite{ji2023deep}.
This result agrees with the observation in~\cite{asnani2022proactive, asnani2022pro_loc}, where proactive-based approaches exhibit improved generalization on manipulation detection and localization tasks.

\minisection{Qualitative Results.}
\cref{fig:cod_visual} visualizes the predicted camouflaged map for DGNet before and after applying \methodName on testing samples of all three datasets. \methodName improves the predicted camouflaged map, with less blurriness along the boundaries and better localization of the camouflaged object. As observed before for GOD, the generated and recovered template has the semantics of the camouflaged objects, which after multiplication intensifies the foreground object, resulting in better segmentation by DGNet.

%============================================================================
%============================================================================
\subsection{Ablation Study}

\minisection{Comparison with Proactive Works.} The prior proactive works perform a different task of image manipulation detection and localization. Therefore, these works are not directly comparable to our proposed proactive wrapper, which performs a different task of object detection as described in Tab.~\ref{tab:rel_works}. However, manipulation localization and COD both involve a prediction of a localization map, segmentation, and fakeness map, respectively. This inspires us to experiment with MaLP~\cite{asnani2022pro_loc} for the task of COD. We train the localization module of MaLP supervised with the COD datasets. The results are shown in Tab.~\ref{tab:baseline_comp_proac}. We see that MaLP is not able to perform well for all three datasets. MaLP is designed for estimating universal templates rather than templates tailored to specific images. %This approach leads to a decline in performance. 
It shows the significance of image-specific templates in object detection. While MaLP's design with image-independent templates is effective for localizing image manipulation, applying it to object detection has a negative impact on performance.

\minisection{Framework Design.} \methodName consists of blocks to improve the object detector.  \cref{tab:abl_framework} ablates different versions of \methodName to highlight the importance of each block in our design. \methodName utilizes an encoder network $\mathcal{E}$ to learn image-dependent templates aiding the detector. 
We remove the encoder $\mathcal{E}$ from our network, replacing it with a fixed template. We observe that the performance deteriorates by a large margin. 
Next, we make this template learnable as proposed in \methodName, but only a single template would be used for all the input images. 
This choice also results in worse performance, highlighting that image-dependent templates are necessary for object detection.
Finally, we remove the decoder network $\mathcal{D}$, which is used to recover the template from the encrypted images. Although this results in a better performance than the pretrained Faster R-CNN, we observe a drop as compared to \methodName. 
Therefore, as discussed in \cref{sec:method}, the recovery of templates is indeed a necessary and beneficial step for boosting the performance of the proactive schemes. 

\minisection{Encryption Process.} \methodName includes an encryption process as described in \cref{eq:trans}, which involves multiplying the template with the input image. This process makes the template act as a mask, highlighting the foreground for better detection. However, prior proactive works~\cite{asnani2022pro_loc, asnani2022proactive} consider adding templates to achieve better results. Thus, we ablate by changing the encryption process to template addition. 
\cref{tab:abl_framework} shows that template addition degrades performance by a significant margin w.r.t.~our multiplication scheme. This shows that encryption is a key step in formulating proactive schemes, and the same encryption process may not work for all tasks. 

%\minisection{Generalization Performance.} 

\minisection{More Training Time.} We perform an ablation to show that the performance gain of the detector is due to our proactive wrapper instead of training for more iterations of the pretrained object detector. Results in \cref{tab:god_more_train} show that although more training iterations for the detector has a performance gain, it's not enough to get the significant margin in performance as achieved by \methodName. This shows that extra training can help, but only up to a certain extent. 

\minisection{Inference Time.} We evaluate the overhead computational cost after applying \methodName on different object detectors are shown in \cref{tab:god_more_train}, averaged across $1,000$ images, on a NVIDIA $V100$ GPU. Our encoder network has $17$ layers, which adds extra cost for inference. For detectors with bulky architectures like Faster R-CNN (ResNet101) and DeTR (transformer), the overhead computational cost is quite small, $8.7\%$ and $7.2\%$, respectively. This additional cost is minor compared to the performance gain of detectors, especially Faster R-CNN. For a lighter detector like YOLOv5, our overhead computational cost increases to $29.1\%$. So, there is a trade-off of applying \methodName to different detectors with varied architectures. \methodName is more beneficial to bulky detectors like two-staged/transformer-based as compared to one-stage detectors. 

%============================================================================
%============================================================================
%============================================================================
\section{Conclusion}
We mathematically prove that the proactive method results in a better-converged model than the passive detector under assumptions and, consequently, a better 2D object detector.
Based on this finding, we propose a proactive scheme wrapper, \methodName, which enhances the performance of camouflaged and generic object detectors. 
The wrapper outputs an image-dependent template using an encoder network, which encrypts the input images. 
These encrypted images are then used to fine-tune the object detector. 
Extensive experiments on MS-COCO, CAMO, \CODTenK, and \NCFourK datasets show that \methodName improves the overall object detection performance for both GOD and COD detectors.
% Our experiments with different camouflaged detectors show that \methodName improves the performance, on CAMO, \CODTenK, and \NCFourK datasets. We show that \methodName also improves the performance of generic object detectors on MS-COCO datasets. 

\minisection{Limitations.}
Our proposed scheme has the following limitations. First, \methodName does not provide a significant gain for recent object detectors such as YOLO and DeTR. Second, the proactive wrapper should be thoroughly tested on other object detectors to show the generalizability of \methodName. Finally, we only experiment with simple multiplication and addition as the encryption scheme. A more sophisticated encryption process might further improve the object detectors' performance. We leave these for our future avenues.

%%%%%%%%% REFERENCES
\clearpage
\nocite{*}
{
\bibliographystyle{ieee_fullname}
\bibliography{egbib}
}

\clearpage
\setcounter{equation}{0}
\setcounter{figure}{0}
\setcounter{table}{0}
\setcounter{section}{0}

\toptitlebar

\begin{center}
    \LARGE \textbf{\textcolor{red}{\methodName}: \textcolor{red}{Pr}oactive \textcolor{red}{Ob}j\textcolor{red}{e}ct \textcolor{red}{D}etection Wrapper} \\ \textbf{-- Supplementary material --}\\
\end{center}

\bottomtitlebar

\section{Proof of Lemma 1}
We begin our proof by considering the image $\vect{i}$ as a column vector and the model as a linear regression model with learnable weights $\vect{w}_t$. The subscript of time $t$ denotes that the weights change as one performs SGD updates.

\minisection{SGD Steps.} We first consider the gradient of weight ($\vect{w}_t$). 
The linear model uses SGD for training, therefore, $\vect{w}_t$ after $t$ gradient steps is given by:
\begin{equation}
    \label{eq:conv_wei}
    \vect{w}_t=\vect{w}_0 - \sum_{i=0}^ts_i\vect{g}_t= \vect{w}_0 - \sum_{i=0}^ts_i\frac{\partial\mathcal{L}}{\partial\vect{w}_t},
\end{equation}
where, for linear regression model with image $\vect{i}$, $\mathcal{L}=f(\vect{w}_t\vect{i}-z)=f(\eta)$. To estimate the gradient $\vect{w}_t$, we have,
\begin{align}
    \label{eq:sgd}
    \vect{g}_t&=\frac{\partial\mathcal{L}(\vect{w}_t\vect{i}-z)}{\partial\vect{w}_t}\nonumber\\
    &=\frac{\partial\mathcal{L}(\vect{w}_t\vect{i}-z)}{\partial(\vect{w}_t\vect{i}-z)}\frac{{\partial(\vect{w}_t\vect{i}-z)}}{\partial\vect{w}_t}\nonumber\\
    &=\frac{\partial\mathcal{L}(\eta)}{\partial\eta}\vect{i}\nonumber\\
    \vect{g}_t&=\vect{i}\upsilon,
\end{align}
where $\upsilon = \frac{\partial\mathcal{L}(\eta)}{\partial\eta}$ is the gradient of the loss function wrt noise.

\minisection{Optimal Weights.} First, we will find the bound of the converged value $\vect{w}_{\infty}$ and the optimal value $\vect{w}_*$. If $\mu_w$ is mean of the learned weight, we have,
\begin{align}
    \mathbb{E}\left(\norm{\vect{w}_{\infty}-\vect{w}_*}_2^2\right)&=\mathbb{E}\left(\norm{\vect{w}_{\infty}-\mu_w+\mu_w-\vect{w}_*}_2^2\right),\nonumber\\
    &=\mathbb{E}((\vect{w}_{\infty}-\mu_w)^T(\vect{w}_{\infty}-\mu_w))+\mathbb{E}((\mu_w-\vect{w}_{*})^T(\mu_w-\vect{w}_{*}))\nonumber\\
    &+2\mathbb{E}((\vect{w}_{\infty}-\mu_w)^T(\mu_w-\vect{w}_{*})),\nonumber\\
    &= \mathbb{E}((\vect{w}_{\infty}-\mu_w)^T(\vect{w}_{\infty}-\mu_w))+\mathbb{E}((\mu_w-\vect{w}_{*})^T(\mu_w-\vect{w}_{*}))
\end{align}

Using $\mathbb{E}(\vect{w}_{\infty}-\mu_w) = \mathbb{E}(\vect{w}_{\infty}) - \mu_w = \mu_w - \mu_w = 0$, we have
\begin{align}
\label{eq:opt_wei}
    \implies \mathbb{E}\left(\norm{\vect{w}_{\infty}-\vect{w}_*}_2^2\right) &=Var(\vect{w}_{\infty})+ \mathbb{E}((\mu_w-\vect{w}_{*})^T(\mu_w-\vect{w}_{*}))
\end{align}
where $Var(\vect{w})=\sum_jw_j^2$.

\minisection{Gradient of Weight.}
Given the image vector $\vect{i}$, and noise $\eta$ are statistically independent, the image and noise gradient $\upsilon$ defined in \cref{eq:sgd} are also statistically independent. We also assume that the distribution of image is normal Gaussian ($\mathbb{E}(\vect{i})=0$). Therefore, the expectation of the gradient $\vect{g}_t$ is given by,
\begin{align}
\label{eq:ex_gra}
    \mathbb{E}(\vect{g}_t)=\mathbb{E}(\vect{i})\mathbb{E}(\upsilon)=0,
\end{align}

Next, the variance of $\vect{g}_t$ is given as
\begin{align}
    Var(\vect{g}_t)&= Var(\vect{i}\upsilon) = \mathbb{E}(\vect{i}^T\vect{i})[Var(\upsilon)+\mathbb{E}^2(\upsilon)] -\mathbb{E}(\vect{i})\mathbb{E}(\upsilon).\nonumber\\
\end{align}
We assume that image pixels are normally distributed. This is common since the networks do a mean subtraction before inputting to the network. Thus, $\mathbb{E}(\vect{i})=0 $. Hence, we have
\begin{align}
\label{eq:var_gra}
    Var(\vect{g}_t)&=\mathbb{E}(\vect{i}^T\vect{i})Var(\upsilon).
\end{align}

\minisection{Converged Weight.}
From \cref{eq:conv_wei}, the expectation of the weight at time $t$ is,
\begin{align}
    \label{eq:exp_conv}
    \mathbb{E}(\vect{w}_t)&=\mathbb{E}(\vect{w}_0)+ \sum_{i=0}^ts_i\mathbb{E}(\vect{g}_j)\nonumber\\
    &=0 \text{     (Using \cref{eq:ex_gra})}\nonumber\\
\end{align}
Therefore, for converged weight,
\begin{align}
    \mathbb{E}(\vect{w}_\infty)&=\lim_{t\to\infty}\mathbb{E}(\vect{w}_t),\nonumber\\
    \mathbb{E}(\vect{w}_\infty)&=\mathbb{E}(\mu_w)=0.
\end{align}

For variance, using \cref{eq:conv_wei} we have, 
\begin{align}
    \label{eq:var_conv}
    Var(\vect{w}_t)&=Var(\vect{w}_0) + (\sum_{i}^ts_j^2)Var(\vect{g}_t).\nonumber\\
    \text{Therefore, we have,}&\nonumber\\
    Var(\vect{w}_\infty)&=\lim_{t\to\infty}(Var(\vect{w}_t))\nonumber\\
    &=Var(\vect{w}_0) +\Big (\lim_{t\to\infty}\sum_{i=}^ts_j^2\Big )Var(\vect{g}_t)\nonumber\\
    Var(\vect{w}_\infty)&=Var(\vect{w}_0) + \mathcal{S'}Var(\vect{g}_t).   
\end{align}

Substituting \cref{eq:var_gra} in the above equation, we have
\begin{equation}
    Var(\vect{w}_\infty)=Var(\vect{w}_0) + \mathcal{S'}\mathbb{E}(\vect{i}^T\vect{i})Var(\upsilon)\\,
\end{equation}
Going back to \cref{eq:opt_wei}, and substituting \cref{eq:exp_conv} and \cref{eq:var_conv}, we have,
\begin{align}
    \label{eq:var_sq}
    \mathbb{E}\left(\norm{\vect{w}_{\infty}-\vect{w}_*}_2^2\right) &=Var(\vect{w}_0) + \mathcal{S'}\mathbb{E}(\vect{i}^T\vect{i})Var(\upsilon)+\mathbb{E}(||\vect{w}_*||^2)\nonumber\\
    \implies \mathbb{E}\left(\norm{\vect{w}_{\infty}-\vect{w}_*}_2^2\right) &=c+\mathcal{S}Var(\upsilon)
\end{align}
where $c$ is independent of loss function $\mathcal{L}$ and $\mathcal{S} = \mathcal{S'}\mathbb{E}(\vect{i}^T\vect{i})$ is also another constant.

\minisection{Lemma $1$.}

We assume that the regression error term $e=\vect{w}^T\vect{i}-\hat{y}$, is drawn from zero mean Gaussian with variance $\sigma^2$ as in \cite{he2019bounding}. So, 
\begin{align}
    Var(\hat{e})&=Var(\vect{w}^T\vect{i}-\hat{y})=\sigma^2.
\end{align}
For a passive detector with converged weights $\vect{w}_{\infty}$, we have,
\begin{align}
    \mathbb{E}\left(\norm{\vect{w}_{\infty}-\vect{w}_*}_2^2\right)&=c+\mathcal{S}Var(\upsilon)\nonumber\\
    &=c+\mathcal{S}Var(e)\nonumber\\
    \implies \mathbb{E}\left(\norm{\vect{w}_{\infty}-\vect{w}_*}_2^2\right)&=c+\mathcal{S}\sigma^2
    \label{eq:passive_converged}
\end{align}

Similarly, for a proactive detector with converged weights $\vect{w}^{'}_{\infty}$, we have
\begin{align}
    \mathbb{E}\left(\norm{\vect{w}^{'}_{\infty}-\vect{w}_*}_2^2\right)&=c+\mathcal{S}Var(\upsilon^{'})
    \label{eq:proactive_converged}
\end{align}

Assume that a proactive detector multiplies the input image vector $\vect{i}$ with a scalar template $s$. 
From \cref{eq:var_sq}, we write the loss term as,
\begin{align}
    \label{eq:pro_loss}
    \mathcal{L}^{'}&= \frac{1}{2}\left(s\vect{w}^T\vect{i}-\hat{y}\right)^2\nonumber\\
    \implies \frac{\partial\mathcal{L}^{'}}{\partial\vect{w}}&=(s\vect{w}^T\vect{i}-\hat{y})s\vect{i}
\end{align}
Taking the variance,
\begin{align}
    Var(\upsilon^{'}) = Var\left(\frac{\partial\mathcal{L}^{'}}{\partial\vect{w}}\right)&=Var((s\vect{w}^T\vect{i}-\hat{y})s\vect{i})\nonumber\\
    &=Var(s(\hat{y}+e)-\hat{y})s^2Var(\vect{i})\nonumber \quad\text{, assuming } \mathbb{E}(\vect{i}) = 0\nonumber\\
    &=Var(s e+(s-1)\hat{y})s^2Var(\vect{i})\nonumber\nonumber\\
    &=(Var(s e)+Var((s-1)\hat{y}))s^2Var(\vect{i})\nonumber\nonumber\\
    &=s^2Var(e)s^2Var(\vect{i}) \quad\quad\quad\quad\text{   , assuming } Var(\hat{y})=0\nonumber\\
    &\le s^2Var(e)s^2 \quad\text{, assuming }Var(\vect{i}) \le 0.5\!\times\!(-1)^2\!+\!0.5\times1^2=1\\
    \implies Var(\upsilon^{'}) &\le s^4\sigma^2
\end{align}
If the magnitude of the scalar template is bounded by $1$ i.e., $s^2 < 1$, we have
\begin{align}
     Var(\upsilon^{'})&<\sigma^2.
\end{align}
The above shows that the gradients in the proactive model has less noise than the passive model (a key for better convergence). Substituting above in \cref{eq:proactive_converged}, we have 
\begin{align}
    \mathbb{E}\left(\norm{\vect{w}^{'}_{\infty}-\vect{w}_*}_2^2\right)&=c+\mathcal{S}Var(\upsilon^{'})\nonumber\\
    &<c+\mathcal{S}\sigma^2\nonumber\\
    &<c+\mathcal{S}Var(\upsilon)\nonumber\\
    \implies \mathbb{E}\left(\norm{\vect{w}^{'}_{\infty}-\vect{w}_*}_2^2\right)&<\mathbb{E}\left(\norm{\vect{w}_{\infty}-\vect{w}_*}_2^2\right).
\end{align}
The last inequality follows trivially from \cref{eq:passive_converged}.

\section{Proof of Theorem 1}
From Lemma $1$, we have,
\begin{align}
    \mathbb{E}\left(\norm{\vect{w}^{'}_{\infty}-\vect{w}_*}_2^2\right)&<\mathbb{E}\left(\norm{\vect{w}_{\infty}-\vect{w}_*}_2^2\right)\nonumber\\
    \implies Var(\vect{w}_\infty^{`})&< Var(\vect{w}_\infty)\nonumber\\
    \implies \mathbb{E}(|\vect{w}_\infty^{`T}\vect{i}-y|)&< \mathbb{E}(|\vect{w}_\infty^T\vect{i}-y|)\nonumber\\
    \implies\mathbb{E}(\hat{y}^{`}-y)&< \mathbb{E}(\hat{y}-y)
\end{align}
Since the proactive detector has a better bounding box prediction,
\begin{align}
    \implies\mathbb{E}(IoU_{2D}^{'})&> \mathbb{E}(IoU_{2D}) 
\end{align}
Since $AP$ is a non-decreasing function of $IoU_{2D}$, we have, 
\begin{align}
    AP^{`}\geq AP.
\end{align}

An important point to note is that the non-decreasing nature does not keep the inequality strict. In other words, we agree that the final AP from passive and pro-active schemes could be equal.  
%we state that AP is a non-decreasing function of IoU. 
However, our experience says that IoU improvements, especially close to $1$, lead to significant AP improvements. Current SoTA detectors already achieve decent IoU; hence, even a slight improvement in IoU improves the AP score.

\clearpage
\section{Implementation Details}
We now include more details of our method here.

\begin{figure}[t!]
\centering
\includegraphics[trim={0 -4 0 0},clip,width=1\textwidth]{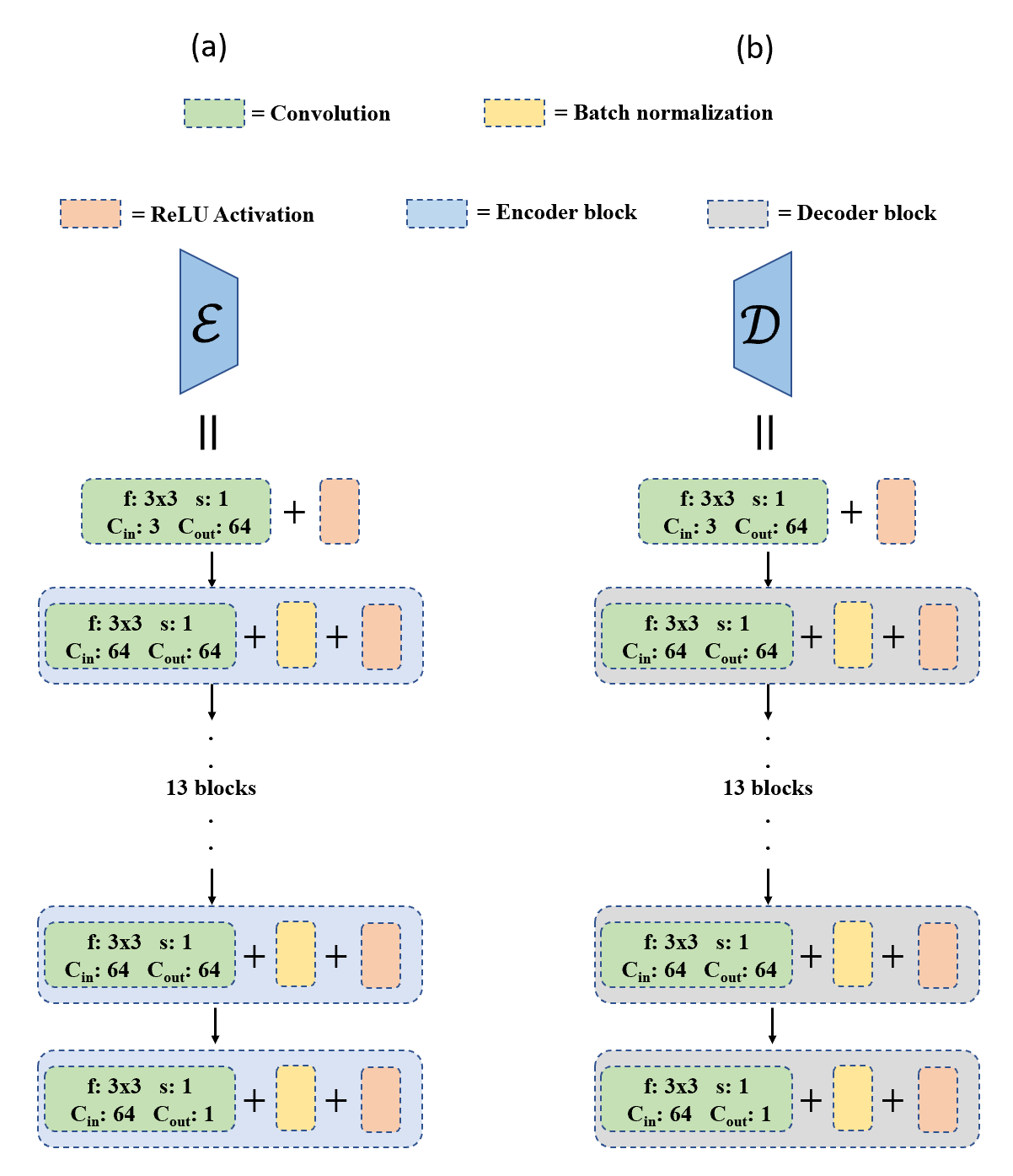}
\caption{\textbf{Architecture} for encoder and decoder network.}
\label{fig:arch}
% \vspace{-3mm}
\end{figure}

\minisection{Network Architecture.} 
The network architecture of encoder $\mathcal{E}$ and decoder $\mathcal{D}$ network used for \methodName is shown in \cref{fig:arch}. Both networks consist of $2$ stem convolution layers and $13$ blocks, each block containing convolutional, batch normalization, and ReLU activation layers. The images are given as input to the encoder network to output the template, which is multiplied by the input images to make them encrypted. The encrypted images are then passed to the decoder network to recover the template. Finally, we input encrypted images to different object detectors to perform detection. 

\minisection{Dataset license information.}
We use benchmark datasets for GOD and COD. The authors for MS-COCO~\cite{lin2014microsoft} dataset specify that the annotations in this dataset, along with this website, belong to the COCO Consortium and are licensed under a Creative Commons Attribution $4.0$ License. The \CODTenK dataset is available for non-commercial purposes only~\cite{fan2021concealed}. The CAMO data is published under the Creative Commons Attribution-NonCommercial-ShareAlike 3.0 License~\cite{le2019anabranch}. Finally, the \NCFourK dataset is available to use for non-commercial purposes. 

\minisection{Experimental Setup and Hyperparameters.} 
\methodName is trained in an end-to-end manner for all the object detectors, with training iterations similar to the pretrained object detector. For both encoder and decoder networks, we use Adam optimizer with a learning rate of $1e^{-5}$. We use different weights of [$\lambda_{OBJ}, \lambda_{E}, \lambda_{D}$] for different object detectors. We use [$7$,$10$,$10$] for Faster-RCNN, [$50$, $1.25$, $4.25$] for YOLOv5, [$50$, $7.5$, $7.5$] for DeTR and [$10$, $0.1$, $0.1$] for DGNet. All experiments are conducted on one NVIDIA A$100$ GPU. 
\iffalse
\begin{table}[t]
%\rowcolors{1}{mygray}{white}
\small
%\begin{center}
\centering
\caption{\textbf{Training weights} for loss functions of different object detectors while training \methodName. } 
\label{tab:weights}
\centering
\begin{adjustbox}{width=0.4\textwidth}
\setlength{\tabcolsep}{0.2cm}
% \rowcolors{2}{mygray}{white}
\begin{tabular}{l|c c c }
\myTopRule
Method & $\lambda_{E}$& $\lambda_{O}$& $\lambda_{OBJ}$\\ 
\myTopRule
Faster-RCNN & & & \\
YOLOv5& & & \\
DeTR& & & \\
DGNet& & & \\
\myTopRule
%\hline
%\hline
\end{tabular}
\end{adjustbox}
%\end{center}
\vspace{-3mm}
\end{table}
\fi

\begin{figure}[t!]
\centering
\includegraphics[trim={0 -4 0 0},clip,width=1\textwidth]{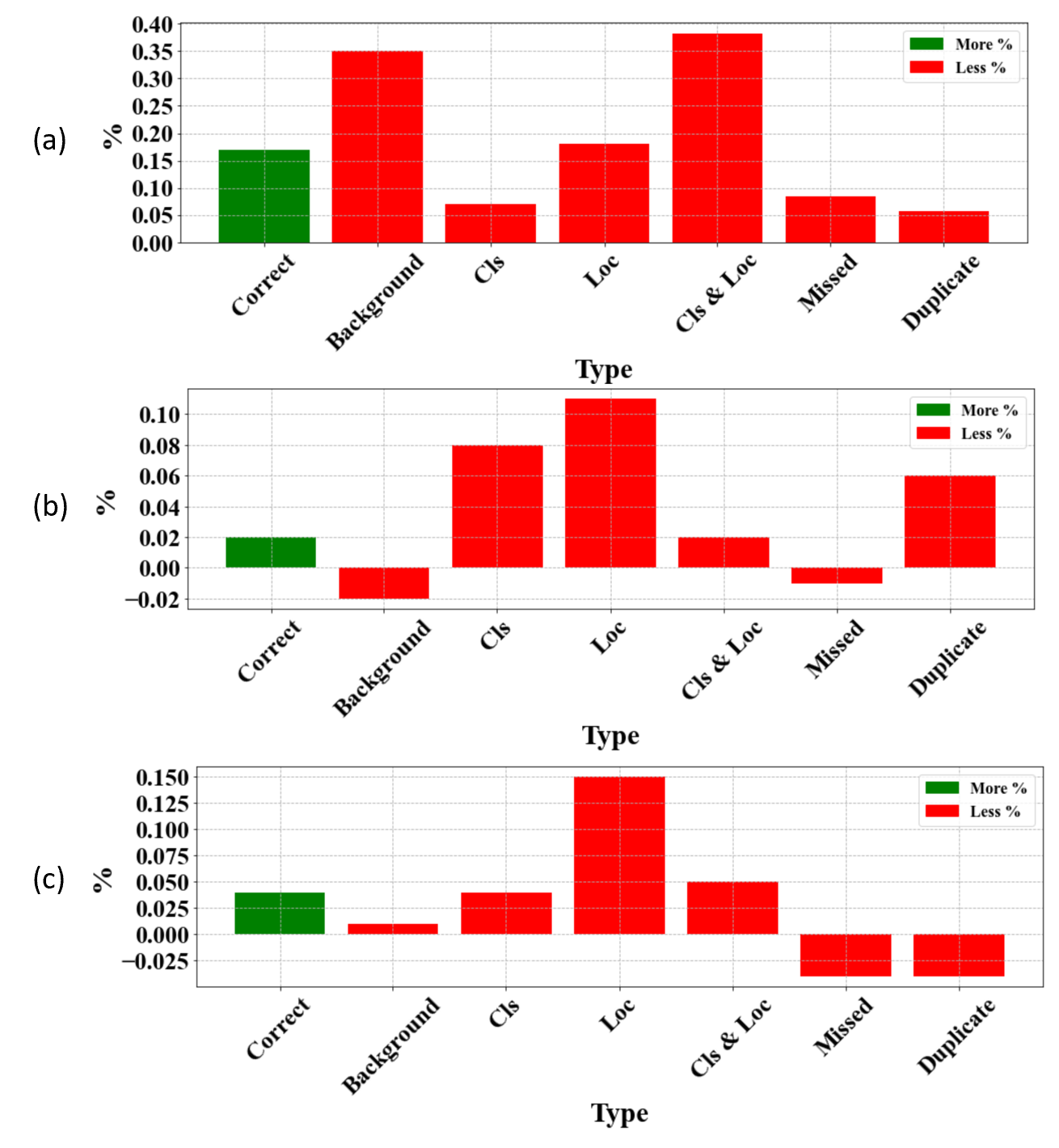}
\caption{\textbf{Error analysis} for (a) Faster-RCNN, (b) YOLOv5, and (c) DeTR. \methodName is able to improve the number of correct predictions and reduce most errors. }
\label{fig:error}
\vspace{-3mm}
\end{figure}

\section{Additional Experiments}
\minisection{Train COD detector DGNet more.}
Similar to the GOD detector, we train the COD detector DGNet for more iterations, similar to after applying \methodName. The results are shown in \cref{tab:cod_more_train}. We see a similar behavior as seen in GOD detectors; the performance improves after training for more iterations, but only up to a certain extent. \methodName is able to improve performance by a larger margin, showing the effectiveness of the proactive schemes.

\begin{table}[t]
%\rowcolors{1}{mygray}{white}
%\small
%\begin{center}
\centering
\caption{\textbf{Ablation of training iterations} on DGNet for more iterations similar to after applying \methodName. } 
\label{tab:cod_more_train}
\centering
\begin{adjustbox}{width=1\textwidth}
\setlength{\tabcolsep}{0.06cm}
% \rowcolors{2}{mygray}{white}
\begin{tabular}{l|c|c c c c| c c c c|c c c c}
\myTopRule
\multirow{2}{*}{Method} & \multirow{2}{*}{Iter} & E$_m\!\uparrow$ & S$_m\!\uparrow$ & wF$_{\beta}\!\uparrow$ & MAE$\downarrow$ & E$_m\!\uparrow$ & S$_m\!\uparrow$ & wF$_{\beta}\!\uparrow$ & MAE$\downarrow$ & E$_m\!\uparrow$ & S$_m\!\uparrow$ & wF$_{\beta}\!\uparrow$ & MAE$\downarrow$ \\ \hhline{~|~|-|-|-|-|-|-|-|-|-|-|-|-}
& & \multicolumn{4}{c|}{CAMO} & \multicolumn{4}{c|}{\CODTenK} & \multicolumn{4}{c}{\NCFourK} \\ \myTopRule
DGNet\cite{ji2023deep} & $1\times$ & $0.859$ & $0.791$ & $0.681$ & $0.079$ & $0.833$ & $0.776$ & $0.603$ & $0.046$ & $0.876$ & $0.815$ & $0.710$ & $0.059$ \\
DGNet\cite{ji2023deep} & $2\times$ & $0.861$ & $0.791$ & $0.682$ & $0.080$ & $0.832$ & $0.778$ & $0.606$ & $0.045$ & $0.875$ & $0.814$ & $0.711$ & $0.059$\\ 
$+$ \methodName & $2\times$ & $\bf0.871$ & $\bf0.797$ & $\bf0.702$ & $\bf0.071$ & $\bf 0.869$ & $\bf0.803$ & $\bf0.661$ & $\bf0.037$ & $\bf 0.900$ & $\bf0.838$ & $\bf0.755$ & $\bf0.049$ \\\hline
\myTopRule
%\hline
%\hline
\end{tabular}
\end{adjustbox}
%\end{center}
\vspace{-3mm}
\end{table}

\minisection{COD loss.}
Our loss design is inspired by the prior proactive works~\cite{asnani2022pro_loc, asnani2022proactive}, which estimate the learnable template by applying a cosine similarity loss. The authors experiment with various loss types, showing the effectiveness of the cosine similarity loss design. However, COD is analogous to the segmentation task, which generally adopts a loss design of cross-entropy loss with dice loss, which might be beneficial for COD. We perform an ablation by applying cross-entropy loss with dice loss for COD. The results are shown in Table~\ref{tab:dice_loss}. We see that our proactive wrapper is not benefiting by removing the cosine similarity loss, proving the study of the prior proactive works.

\begin{table}[!t]
% \small
%\begin{center}
\caption{Ablation of dice loss with cross-entropy (CE) loss \vs{} cosine similarity} 
\label{tab:dice_loss}
\begin{adjustbox}{width=1\textwidth}
\centering
\setlength{\tabcolsep}{0.05cm}
\rowcolors{4}{mygray}{white}
\begin{tabular}{l|c c c c| c c c c|c c c c}
\myTopRule
%\hline
%\rowcolor{mygray}
& \multicolumn{4}{c|}{CAMO} & \multicolumn{4}{c|}{\CODTenK} & \multicolumn{4}{c}{\NCFourK}\\ \hhline{~|-|-|-|-|-|-|-|-|-|-|-|-}
%\rowcolor{mygray}
\multirow{-2}{*}{Method} & E$_m\!\uparrow$ & S$_m\!\uparrow$ & wF$_{\beta}\!\uparrow$ & MAE$\downarrow$ & E$_m\!\uparrow$ & S$_m\!\uparrow$ & wF$_{\beta}\!\uparrow$ & MAE$\downarrow$ & E$_m\!\uparrow$ & S$_m\!\uparrow$ & wF$_{\beta}\!\uparrow$ & MAE$\downarrow$\\
\myTopRule
Dice $+$ CE loss & $0.831$ & $0.782$ & $0.688$ & $0.084$ & $0.810$ & $0.795$ & $0.646$ & $0.045$ & $0.874$ & $0.817$ & $0.721$ & $0.060$\\
Cosine similarity & $\bf0.871$ & $\bf0.797$ & $\bf0.702$ & $\bf0.071$ & $\bf 0.869$ & $\bf0.803$ & $\bf0.661$ & $\bf0.037$ & $\bf 0.900$ & $\bf0.838$ & $\bf0.755$ & $\bf0.049$\\
\myTopRule
%\hline
%\hline
\end{tabular}
\end{adjustbox}
%\end{center}
\vspace{-2mm}
\end{table}

\minisection{Error analysis.}
Following~\cite{bolya2020tide}, there can be a number of errors that deteriorate the performance of the object detector. These are:
\begin{enumerate}
\item Classification error (Cls): Localized correctly but classified incorrectly.
\item Localization error (Loc): Classified correctly but localized incorrectly.
\item Both Classification and Localization error (Cls \& Loc): Classified and localized incorrectly.
\item Duplicate detection error (Duplicate): Would be correct if not for a higher scoring detection.
\item Background error (Background): Detected background as foreground.
\item Missed target error (Missed): All undetected targets \ie false negatives, which are not already covered by classification or localization errors.
\end{enumerate}
\cref{fig:error} shows the error analysis for three object detectors, namely, Faster-RCNN, YOLOv5, and DeTR. \methodName improves the number of correct predictions of all three detectors, especially for Faster-RCNN, where the number of correct predictions increases by around $17\%$. For DeTR and YOLOv5, the improvement is less, which is evident from the less increase in correct predictions. The major improvement for all three detectors comes from classification and localization-related errors. All these errors decrease after \methodName is applied to all the detectors. Further, Faster-RCNN, being an old detector, makes a lot of background errors, which are reduced by a significant margin after applying \methodName. The gain is not much for DeTR and YOLOv5, which tend to make fewer background errors. Finally, one-stage detectors suffer mostly from the problem of duplicate detection, which is remedied by the \methodName.

\section{Potential Negative Societal Impact}
\methodName utilizes a proactive scheme to benefit object detection. Our approach can be considered a benign adversarial attack on object detectors. However, with a change in the objective function, \methodName could also be used as an adversarial attack to deteriorate the performance of different object detectors. This might pose a threat to object detectors, whether used for GOD or COD, and some forms of adversarial training might be required to prevent the threat of adversarial attacks.

\end{document}